\documentclass{ecai} 

\usepackage{microtype}
\usepackage{graphicx}
\usepackage{subfigure}
\usepackage{booktabs}

\usepackage{amsmath}
\usepackage{amssymb}
\usepackage{mathtools}
\usepackage{amsthm}
\usepackage{xspace}
\usepackage{array}
\usepackage{arydshln}
\usepackage{orcidlink}
\usepackage{cleveref}
\usepackage{multirow} 

\usepackage{tikz}
\usepackage{pgfplots}
\usepackage{rotating}
\pgfplotsset{width=7cm,compat=1.18}
\usetikzlibrary{shapes,arrows.meta,3d,shapes.geometric,calc,shadows,matrix,positioning,patterns}
\usepackage{listofitems}
\usepackage{enumitem}
\setlist[itemize]{leftmargin=5.5mm}

\colorlet{myred}{red!80!black}
\colorlet{myblue}{blue!80!black}
\colorlet{mygreen}{green!60!black}
\colorlet{mydarkred}{myred!40!black}
\colorlet{mydarkblue}{myblue!40!black}
\colorlet{mydarkgreen}{mygreen!40!black}

\definecolor{lightblue}{rgb}{0.145,0.6666,1}
\definecolor{darkblue}{rgb}{0.169,0.549,0.745}
\definecolor{darkgreen}{rgb}{0.015,0.6215,0.4413}

\definecolor{myorange}{RGB}{250, 184, 2}
\definecolor{mypink}{RGB}{247, 10, 125}
\definecolor{myblue}{RGB}{2, 106, 250}
\definecolor{mypurple}{RGB}{147, 2, 250}

\DeclareMathOperator*{\argmax}{arg\,max}
\DeclareMathOperator*{\argmin}{arg\,min}

\graphicspath{{./figures}}

\newcommand{\BibTeX}{B\kern-.05em{\sc i\kern-.025em b}\kern-.08em\TeX}

\begin{document}


\begin{frontmatter}


\paperid{1675} 


\title{Be Persistent: Towards a Unified Solution for Mitigating Shortcuts in Deep Learning}


\author{\fnms{Hadi}~\snm{M.~Dolatabadi}\textsuperscript{\orcidlink{0000-0001-9418-1487}}}
\author{\fnms{Sarah}~\snm{M.~Erfani}\textsuperscript{\orcidlink{0000-0003-0885-0643}}\thanks{Corresponding Author. Email: sarah.erfani@unimelb.edu.au}}
\author{\fnms{Christopher}~\snm{Leckie}\textsuperscript{\orcidlink{0000-0002-4388-0517}}}

\address{School of Computing and Information Systems, The University of Melbourne, Parkville, Victoria, Australia.}


\begin{abstract}
Deep neural networks~(DNNs) are vulnerable to shortcut learning: rather than learning the intended task, they tend to draw inconclusive relationships between their inputs and outputs.
Shortcut learning is ubiquitous among many failure cases of neural networks, and traces of this phenomenon can be seen in their generalizability issues, domain shift, adversarial vulnerability, and even bias towards majority groups.
In this paper, we argue that this commonality in the cause of various DNN issues creates a significant opportunity that should be leveraged to find a unified solution for shortcut learning.
To this end, we outline the recent advances in topological data analysis~(TDA), and persistent homology~(PH) in particular, to sketch a unified roadmap for detecting shortcuts in deep learning.
We demonstrate our arguments by investigating the topological features of computational graphs in DNNs using two cases of unlearnable examples and bias in decision-making as our test studies.
Our analysis of these two failure cases of DNNs reveals that finding a unified solution for shortcut learning in DNNs is not out of reach, and TDA can play a significant role in forming such a framework.
\end{abstract}

\end{frontmatter}


\section{Introduction}
\label{sec:introduction}

Deep neural networks~(DNNs) have reshaped the means of data processing and generation in the past decade.
Many of the recent advances in different areas of machine learning, whether in natural language processing~\citep{vaswani2017attention, radford2018improving} or computer vision~\citep{he2016deep, song2019generative}, can be attributed to the representational power of DNNs.
For instance, it has been shown that neural networks can surpass human-level accuracy in object detection~\citep{russakovsky2015imagenet}.

In contrast to humans, DNNs notoriously rely on spurious features and draw meaningless conclusions during decision-making~\citep{beery2018cows, singala2022salient, kirichenko2023spurious}.
For instance, consider a DNN trained over a dataset that always depicts cows on the grass.
Such a model considers grass as a spurious feature, and only detects cows where the background includes greenery (see~\Cref{fig:cows}).
This simple example indicates a serious flaw in DNNs' decision-making, rendering it dangerous for sensitive applications to be at the mercy of neural networks~\citep{wexler2017computer}. 

    \newsavebox\dnncomplicated
    \begin{lrbox}{\dnncomplicated}
 	\tikzstyle{node}=[very thick,circle,draw=myblue,minimum size=22,inner sep=0.5,outer sep=0.6]
    \tikzstyle{connect}=[-,thick]
    \tikzset{ 
      node 1/.style={node,black,draw=darkgreen,fill=black!15},
      node 2/.style={node,black,draw=blue,fill=black!15},
      node 3/.style={node,black,draw=red,fill=black!15},
    }
    \def\nstyle{int(\lay<\Nnodlen?min(2,\lay):3)}
    \resizebox{0.30\textwidth}{!}{
    \begin{tikzpicture}[x=2.4cm,y=1.2cm]
      \draw [dashed, fill=blue!20] (0.65,-3.45) rectangle (4.35,1.90);
    
      \readlist\Nnod{4,5,4,3} 
      \readlist\Nstr{n,m_1,m_2,k} 
      \readlist\Cstr{x,h^{(\prev)}, y} 
      \def\yshift{0.55} 
      
      \foreachitem \N \in \Nnod{
        \def\lay{\Ncnt} 
        \pgfmathsetmacro\prev{int(\Ncnt-1)} 
        \foreach \i [evaluate={\c=int(\i==\N); \y=\N/2-\i-\c*\yshift;
                     \x=\lay; \n=\nstyle;
                     \index=(\i<\N?int(\i):"\Nstr[\Ncnt]");}] in {1,...,\N}{ 
          \node[node \n] (N\lay-\i) at (\x,\y) {};
          
          \ifnumcomp{\lay}{>}{1}{ 
            \foreach \j in {1,...,\Nnod[\prev]}{ 
              \draw[white,line width=1.2,shorten >=1] (N\prev-\j) -- (N\lay-\i);
              \draw[connect] (N\prev-\j) -- (N\lay-\i);
            }
            \ifnum \lay=\Nnodlen
              \draw[connect] (N\lay-\i) --++ (0.5,0); 
            \fi
          }{
            \draw[connect] (0.5,\y) -- (N\lay-\i); 
          }
          
        }
        \path (N\lay-\N) --++ (0,1+\yshift) node[midway,scale=1.6] {$\vdots$}; 
      }
      \draw[black,line width=3.25,shorten >= 1] (N1-2) -- (N2-2);
      \draw[black,line width=2.25,shorten >= 1] (N1-2) -- (N2-3);
      \draw[black,line width=3.25,shorten >= 1] (N2-2) -- (N3-1);
      \draw[black,line width=2.75,shorten >= 1] (N2-3) -- (N3-2);
      \draw[black,line width=3.25,shorten >= 1] (N2-3) -- (N3-4);
      \draw[black,line width=2.50,shorten >= 1] (N3-1) -- (N4-2);
      \draw[black,line width=3.25,shorten >= 1] (N3-2) -- (N4-2);
      \draw[black,line width=2.75,shorten >= 1] (N3-4) -- (N4-2);
      
    \end{tikzpicture}}
    \end{lrbox}
    \newsavebox\dnnshort
    \begin{lrbox}{\dnnshort}
 	\tikzstyle{node}=[very thick,circle,draw=myblue,minimum size=22,inner sep=0.5,outer sep=0.6]
    \tikzstyle{connect}=[-,thick]
    \tikzset{ 
      node 1/.style={node,black,draw=darkgreen,fill=black!15},
      node 2/.style={node,black,draw=blue,fill=black!15},
      node 3/.style={node,black,draw=red,fill=black!15},
    }
    \def\nstyle{int(\lay<\Nnodlen?min(2,\lay):3)}
    \resizebox{0.30\textwidth}{!}{
    \begin{tikzpicture}[x=2.4cm,y=1.2cm]
      \draw [dashed, fill=blue!20] (0.65,-3.45) rectangle (4.35,1.90);
    
      \readlist\Nnod{4,5,4,3} 
      \readlist\Nstr{n,m_1,m_2,k} 
      \readlist\Cstr{x,h^{(\prev)}, y} 
      \def\yshift{0.55} 
      
      \foreachitem \N \in \Nnod{
        \def\lay{\Ncnt} 
        \pgfmathsetmacro\prev{int(\Ncnt-1)} 
        \foreach \i [evaluate={\c=int(\i==\N); \y=\N/2-\i-\c*\yshift;
                     \x=\lay; \n=\nstyle;
                     \index=(\i<\N?int(\i):"\Nstr[\Ncnt]");}] in {1,...,\N}{ 
          \node[node \n] (N\lay-\i) at (\x,\y) {};
          
          \ifnumcomp{\lay}{>}{1}{ 
            \foreach \j in {1,...,\Nnod[\prev]}{ 
              \draw[white,line width=1.2,shorten >=1] (N\prev-\j) -- (N\lay-\i);
              \draw[connect] (N\prev-\j) -- (N\lay-\i);
            }
            \ifnum \lay=\Nnodlen
              \draw[connect] (N\lay-\i) --++ (0.5,0); 
            \fi
          }{
            \draw[connect] (0.5,\y) -- (N\lay-\i); 
          }
          
        }
        \path (N\lay-\N) --++ (0,1+\yshift) node[midway,scale=1.6] {$\vdots$}; 
      }
      \draw[black,line width=3.25,shorten >= 1] (N1-1) -- (N2-2);
      \draw[black,line width=3.25,shorten >= 1] (N2-2) -- (N3-3);
      \draw[black,line width=3.25,shorten >= 1] (N3-3) -- (N4-3);
      
    \end{tikzpicture}}
    \end{lrbox}
    
\begin{figure}[tb!]
        \centering
     	\resizebox{0.375\textwidth}{!}{
			\begin{tikzpicture}[node distance=1cm, auto]
			\node [inner sep=0pt] at (0.5,2) (Inp) {\includegraphics[width=1.75cm,height=1.75cm]{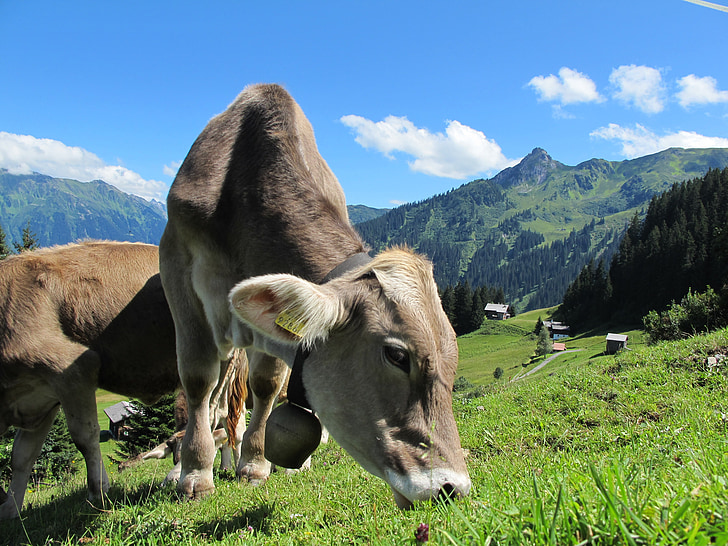}};
			\node [inner sep=0pt] at (0.5,0) (Adv) {\includegraphics[width=1.75cm,height=1.75cm]{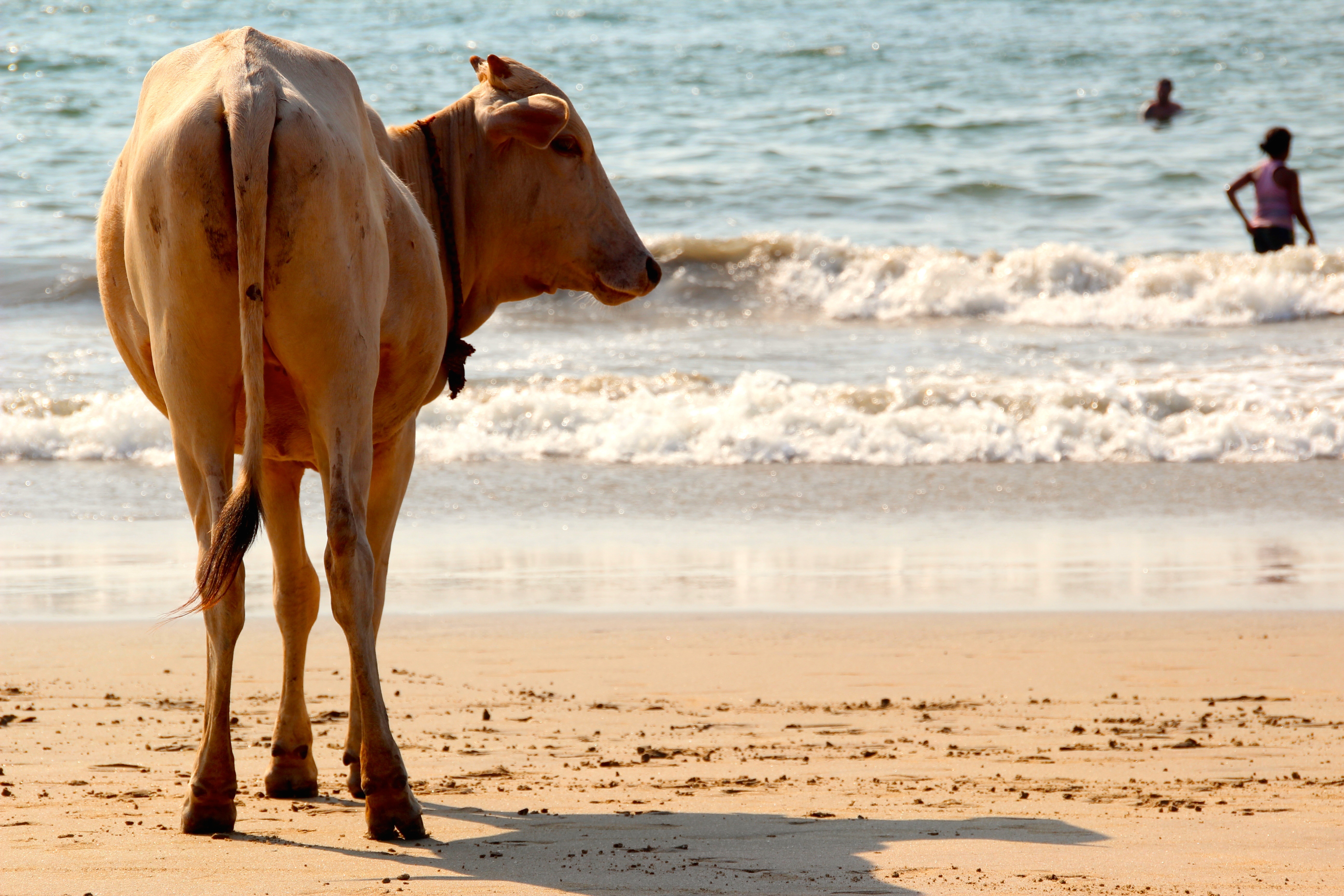}};
			
			\draw[-{Latex[length=3mm, width=1.25mm]}, line width=0.4mm] (1.45, 2) -- (2.15, 2);
            \draw[-{Latex[length=3mm, width=1.25mm]}, line width=0.4mm] (1.45, 0) -- (2.15, 0);
			
			\node [inner sep=0pt, scale=0.4] at (3.2,2) (DNN) {\usebox\dnnshort};
            \node [inner sep=0pt, scale=0.4] at (3.2,0) (DNN) {\usebox\dnncomplicated};

            \draw[-{Latex[length=3mm, width=1.25mm]}, line width=0.4mm] (4.50, 2) -- (5.20, 2);
            \draw[-{Latex[length=3mm, width=1.25mm]}, line width=0.4mm] (4.50, 0) -- (5.20, 0);

            \node [inner sep=0pt, rotate=90] at (5.5, 2) (mul) {Cow!};
			\node [inner sep=0pt, rotate=90] at (5.5, 0) (mul) {Beach!};
			
			\node [inner sep=0pt, rotate=90] at (5.5,2.65) (good) {\includegraphics[width=0.35cm,height=0.35cm]{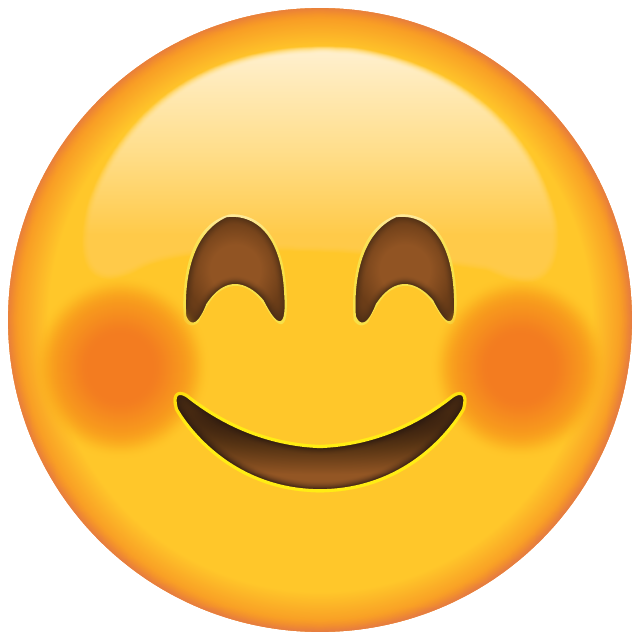}};
			\node [inner sep=0pt, rotate=90] at (5.5,0.75) (bad) {\includegraphics[width=0.35cm,height=0.35cm]{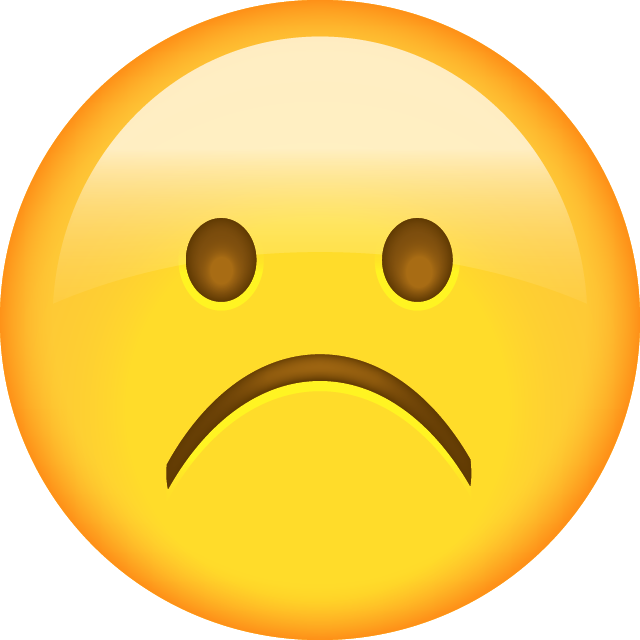}};
   
	    \end{tikzpicture}}
     \caption{\label{fig:cows}Neural networks exhibit spurious correlations, which is a special case of shortcut learning. For instance, DNN trained over data that depicted cows exclusively over grass has created a spurious correlation where cows are only detected in the presence of grass~\citep{beery2018cows}.}
     \vspace{0.25in}
\end{figure}
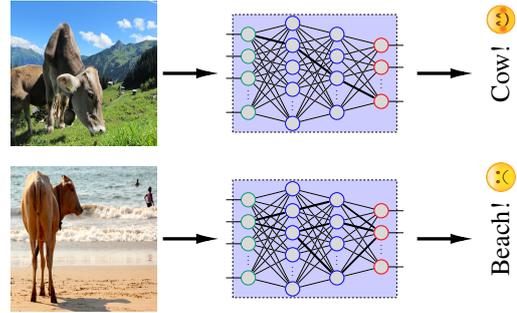

Spurious correlations are argued to be the result of a broader phenomenon known as \textit{shortcut learning}~\citep{geirhos2020shortcut}.
Shortcuts are unintended decision rules that are learned by neural networks~(e.g., detecting grass instead of cows).
These shortcuts are argued to be the result of following the ``Principle of Least Effort''~\citep{zipf2016human} in which learning the unintended solution takes less effort compared to the actual task~\citep{geirhos2020shortcut}~(e.g., detecting greenery would be much easier than detecting cows in different colors and shapes).
Various failure cases of neural networks, from domain shift~\citep{wang2018domain} and adversarial examples~\citep{szegedy2014intriguing} to even bias and fairness~\citep{hardt2023fairness}, can be traced back to shortcut learning~\citep{geirhos2020shortcut}.
This surprising alignment in the cause of these failure cases is a great opportunity that should be leveraged towards finding a unified solution.

In this paper, we argue that topological data analysis~(TDA)~\citep{carlsson2009topology} can play an invaluable role in detecting and mitigating shortcuts in deep learning.
In particular, we demonstrate that shortcut learning leaves tractable paths in the computational graph of a neural network.
To unveil shortcuts in such a high-dimensional space containing thousands of neurons, we leverage the representational power of persistent homology~(PH)~\citep{edelsbrunner2008persistent} in revealing connected components between neurons.
We empirically show that neural networks that learn such shortcuts leave a signature in the topology of their computational graph.
Our experiments reveal that these features are statistically different from those of benign neural networks. 

To demonstrate our point, we choose two failure cases of DNNs in down-stream tasks: unlearnable examples~\citep{huang2021emn, yu2022shr} and bias~\citep{buolamwini2018gender}.
Unlearnable examples are imperceptible perturbations added to the training data to prevent a DNN from learning meaningful patterns from the data~\citep{huang2021emn}.
On the other hand, bias in neural networks can happen when a DNN misuses certain sensitive attributes from the data (such as skin color) to make its final decision~\citep{buolamwini2018gender}.
As seen, the origins of these two issues are quite different.
However, we show that TDA can provide a unified means to shed light on both misbehaviors in DNNs.
Our promising findings call for more research into the topological analysis of neural networks for treating shortcuts, with the hope that such a solution could mitigate various failures of DNNs once and for all. 

Our contributions can be summarized as follows:
\vspace{-0.75em}
\begin{itemize}\setlength\itemsep{0.25em}
\item We target shortcut learning as the main culprit among many failure cases of neural networks~\citep{geirhos2020shortcut} and argue that the research community should pay more attention to persistent homology to mitigate this issue.
To the best of our knowledge, this is the first paper that proposes a unified roadmap for mitigating shortcut learning.

\item We use two emerging issues in neural networks, namely unlearnable examples and bias in decision-making, as our case-studies to demonstrate the applicability of the same persistent homology framework to completely different issues in neural networks. Training more than a dozen models in each case, our experimental results demonstrate that persistent homology can easily reveal the differences between benign and affected models.

\item Finally, we pinpoint some of the most important future research directions in this area that could lead to a unified solution for shortcut learning using persistent homology.

\end{itemize}

\section{Background}
\label{sec:background}
This section reviews the background related to persistent homology.
Rather than an in-depth mathematical discussion, we focus on covering the intuition and a general understanding of persistent homology.
We refer the interested reader to~\citet{edelsbrunner2010persistent} for a formal introduction to the topics discussed here. 

\subsection{Simplicial Homology}
\label{background:simplicial}
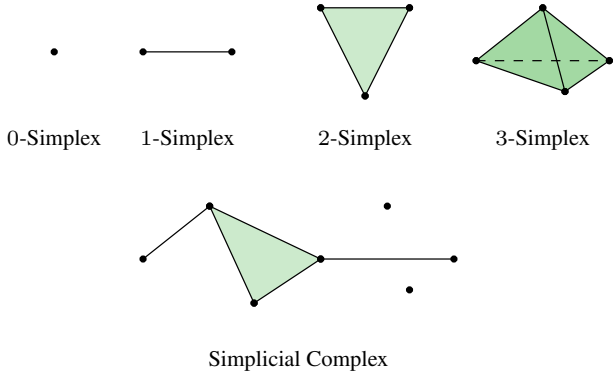
\begin{figure}[tb!]
        \centering
     	\resizebox{0.45\textwidth}{!}{
        \begin{tikzpicture}
            \tikzstyle{point}=[circle,thick,draw=black,fill=black,inner sep=0pt,minimum width=1.5pt,minimum height=1.5pt]
            \node (a)[point] at (0,0) {};
            \node [inner sep=0pt] at (0, -1.0) (txt) {\scriptsize $0$-Simplex};

            \node (b)[point] at (1,0) {};
            \node (c)[point] at (2,0) {};
            \draw (b.center) -- (c.center);
            \node [inner sep=0pt] at (1.5, -1.0) (txt) {\scriptsize $1$-Simplex};

            \node (d)[point] at (3.00, 0.5) {};
            \node (e)[point] at (3.50,-0.5) {};
            \node (f)[point] at (4.00, 0.5) {};
            \draw[fill=mygreen, fill opacity=0.2] (d.center) -- (e.center) -- (f.center) -- cycle;
            \node (d)[point] at (3.00, 0.5) {};
            \node (e)[point] at (3.50,-0.5) {};
            \node (f)[point] at (4.00, 0.5) {};
            \node [inner sep=0pt] at (3.5, -1.0) (txt) {\scriptsize $2$-Simplex};

            \node (g)[point] at (4.75,-0.10) {};
            \node (h)[point] at (5.50, 0.50) {};
            \node (i)[point] at (6.25,-0.10) {};
            \node (j)[point] at (5.75,-0.45) {};
            \path[fill=mygreen, fill opacity=0.2] (g.center) -- (h.center) -- (i.center) -- cycle;
            \path[fill=mygreen, fill opacity=0.2] (g.center) -- (h.center) -- (j.center) -- cycle;
            \path[fill=mygreen, fill opacity=0.2] (g.center) -- (i.center) -- (j.center) -- cycle;
            \path[fill=mygreen, fill opacity=0.2] (h.center) -- (i.center) -- (j.center) -- cycle;
            \node (g)[point] at (4.75,-0.10) {};
            \node (h)[point] at (5.50, 0.50) {};
            \node (i)[point] at (6.25,-0.10) {};
            \node (j)[point] at (5.75,-0.45) {};
            \draw (g.center) -- (h.center);
            \draw (i.center) -- (h.center);
            \draw (j.center) -- (h.center);
            \draw (j.center) -- (i.center);
            \draw (g.center) -- (j.center);
            \draw [dashed] (g.center) -- (i.center);
            \node [inner sep=0pt] at (5.5, -1.0) (txt) {\scriptsize $3$-Simplex};

            \begin{scope}[xshift=-0.5cm, yshift=-2.25cm]
            \node (k)[point] at (3.50,-0.10) {};
            \node (l)[point] at (4.25, 0.50) {};
            \node (m)[point] at (5.00,-0.10) {};
            \node (n)[point] at (4.50,-0.45) {};
            \node (o)[point] at (1.50,-0.10) {};
            \node (p)[point] at (2.25, 0.50) {};
            \node (q)[point] at (2.75,-0.60) {};
            \end{scope}
            \draw (p.center) -- (o.center);
            \draw[fill=mygreen, fill opacity=0.2] (p.center) -- (q.center) -- (k.center) -- cycle;
            \begin{scope}[xshift=-0.5cm, yshift=-2.25cm]
            \node (k)[point] at (3.50,-0.10) {};
            \node (p)[point] at (2.25, 0.50) {};
            \node (q)[point] at (2.75,-0.60) {};
            \end{scope}
            \draw (k.center) -- (m.center);
            \node [inner sep=0pt] at (2.75, -3.5) (txt) {\scriptsize Simplicial Complex};
   
	    \end{tikzpicture}}
     \caption{\label{fig:simplicial} A $k$-Simplex can be regarded as the convex hull of $k+1$ points. A simplicial complex is a union of such simplices.}
     \vspace{0.25in}
     \end{figure}

The field of computational topology~\citep{edelsbrunner2010persistent} is concerned with designing the mathematical tools required for finding topological features of low- and high-dimensional manifolds.
To this end, the notion of \textit{homology groups} in a \textit{simplicial complex} is used.
Informally speaking, a simplicial complex is a discretized graph representation of the data that contains information about nodes, edges, triangles, and their higher order equivalents~(see~\Cref{fig:simplicial}).
A homology group of rank $d$, denoted by $\mathcal{H}_d$, is a $d$-dimensional topological descriptor for simplicial complexes.
In this paper, we mainly work with $0$- and $1$-dimensional homology groups, which encode the number of connected components and cycles/loops in a simplicial complex.

\subsection{Persistent Homology}
\label{sec:background:persistent}
\textit{Persistent homology}~(PH)~\citep{barannikov1994framed, edelsbrunner2008persistent} extends the notion of simplicial homology to data observations.
Rather than treating the data as samples from their true underlying manifold, PH aims to identify the most \textit{persistent} topological structures within the data points, assuming noisy observations.
To this end, the evolution of topological features at multiple granularities using a \textit{filtration} is observed.
A filtration creates a sequence of simplicial complexes at growing granularities, which enables us to track the evolution of topological features between consecutive levels of granularity.
Each topological structure is created at a particular granularity and destroyed at another one.
We refer to these as the birth and death times of our topological features, respectively.
We can then collect all these times as points in a $2$-dimensional plane, creating a so-called \textit{persistent diagram}~(PD).
Naturally, we would be interested in structures that have a long life-time/persistence (which is the subtraction of the death and birth time).
This is because such structures are unlikely to have been generated by the noisy observations, and as such, are representing the true data manifold.

\subsection{Vietoris-Rips Filtration}
\label{sec:background:vietoris}
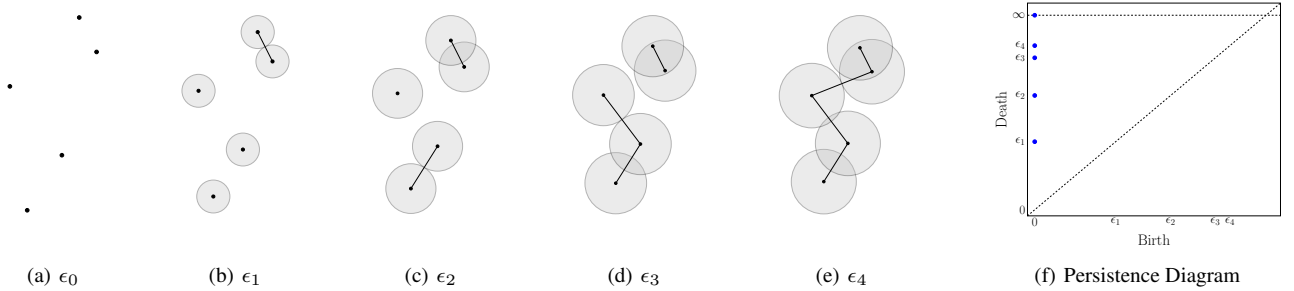
\begin{figure*}[tb!]
        \centering
        \hfill
        \subfigure[$\epsilon_0$\label{fig:vr:e0}]{\raisebox{0.15in}{\rotatebox{90}{
        \resizebox{0.15\textwidth}{!}{
        \begin{tikzpicture}
            \tikzstyle{point}=[circle,draw=black,fill=black,inner sep=0pt,minimum width=1.5pt,minimum height=1.5pt]

            \node (b)[point] at (0.70,-0.00) {};
            \node (c)[point] at (1.50,-0.50) {};
            \node (e)[point] at (2.50, 0.25) {};
            \node (f)[point] at (3.00,-1.00) {};
            \node (g)[point] at (3.50,-0.75) {};
   
    	    \end{tikzpicture}}}}}\hfill
        \subfigure[$\epsilon_1$\label{fig:vr:e1}]{\raisebox{0.15in}{\rotatebox{90}{
        \resizebox{0.15\textwidth}{!}{
        \begin{tikzpicture}
            \tikzstyle{point}=[circle,draw=black,fill=black,inner sep=0pt,minimum width=1.5pt,minimum height=1.5pt]

            \node[circle,draw,fill=lightgray, inner sep=0pt, minimum size=0.56cm, opacity=.3] (b) at (0.7,-0.000){};
            \node[circle,draw,fill=lightgray, inner sep=0pt, minimum size=0.56cm, opacity=.3] (c) at (1.50,-0.50){};
            \node[circle,draw,fill=lightgray, inner sep=0pt, minimum size=0.56cm, opacity=.3] (d) at (2.50, 0.25){};
            \node[circle,draw,fill=lightgray, inner sep=0pt, minimum size=0.56cm, opacity=.3] (e) at (3.00,-1.00){};
            \node[circle,draw,fill=lightgray, inner sep=0pt, minimum size=0.56cm, opacity=.3] (f) at (3.50,-0.75){};
            
            \node (b)[point] at (0.70,-0.000) {};
            \node (c)[point] at (1.50,-0.50) {};
            \node (e)[point] at (2.50, 0.25) {};
            \node (f)[point] at (3.00,-1.0) {};
            \node (g)[point] at (3.50,-0.75) {};

            \draw (g.center) -- (f.center);
            
    	    \end{tikzpicture}}}}}\hfill
        \subfigure[$\epsilon_2$\label{fig:vr:e2}]{\raisebox{0.15in}{\rotatebox{90}{
        \resizebox{0.15\textwidth}{!}{
        \begin{tikzpicture}
            \tikzstyle{point}=[circle,draw=black,fill=black,inner sep=0pt,minimum width=1.5pt,minimum height=1.5pt]

            \node[circle,draw,fill=lightgray, inner sep=0pt, minimum size=0.94cm, opacity=.3] (b) at (0.7,-0.000){};
            \node[circle,draw,fill=lightgray, inner sep=0pt, minimum size=0.94cm, opacity=.3] (c) at (1.50,-0.50){};
            \node[circle,draw,fill=lightgray, inner sep=0pt, minimum size=0.94cm, opacity=.3] (d) at (2.50, 0.25){};
            \node[circle,draw,fill=lightgray, inner sep=0pt, minimum size=0.94cm, opacity=.3] (e) at (3.00,-1.00){};
            \node[circle,draw,fill=lightgray, inner sep=0pt, minimum size=0.94cm, opacity=.3] (f) at (3.50,-0.75){};
            
            \node (b)[point] at (0.70,-0.000) {};
            \node (c)[point] at (1.50,-0.50) {};
            \node (e)[point] at (2.50, 0.25) {};
            \node (f)[point] at (3.00,-1.0) {};
            \node (g)[point] at (3.50,-0.75) {};

            \draw (g.center) -- (f.center);
            \draw (b.center) -- (c.center);
            
    	    \end{tikzpicture}}}}}\hfill
        \subfigure[$\epsilon_3$\label{fig:vr:e3}]{\raisebox{0.15in}{\rotatebox{90}{
        \resizebox{0.15\textwidth}{!}{
        \begin{tikzpicture}
            \tikzstyle{point}=[circle,draw=black,fill=black,inner sep=0pt,minimum width=1.5pt,minimum height=1.5pt]

            \node[circle,draw,fill=lightgray, inner sep=0pt, minimum size=1.25cm, opacity=.3] (b) at (0.7,-0.000){};
            \node[circle,draw,fill=lightgray, inner sep=0pt, minimum size=1.25cm, opacity=.3] (c) at (1.50,-0.50){};
            \node[circle,draw,fill=lightgray, inner sep=0pt, minimum size=1.25cm, opacity=.3] (d) at (2.50, 0.25){};
            \node[circle,draw,fill=lightgray, inner sep=0pt, minimum size=1.25cm, opacity=.3] (e) at (3.00,-1.00){};
            \node[circle,draw,fill=lightgray, inner sep=0pt, minimum size=1.25cm, opacity=.3] (f) at (3.50,-0.75){};
            
            \node (b)[point] at (0.70,-0.000) {};
            \node (c)[point] at (1.50,-0.50) {};
            \node (e)[point] at (2.50, 0.25) {};
            \node (f)[point] at (3.00,-1.0) {};
            \node (g)[point] at (3.50,-0.75) {};

            \draw (g.center) -- (f.center);
            \draw (b.center) -- (c.center);
            \draw (c.center) -- (d.center);
            
    	\end{tikzpicture}}}}}\hfill
        \subfigure[$\epsilon_4$\label{fig:vr:e4}]{\raisebox{0.15in}{\rotatebox{90}{
        \resizebox{0.15\textwidth}{!}{
        \begin{tikzpicture}
            \tikzstyle{point}=[circle,draw=black,fill=black,inner sep=0pt,minimum width=1.5pt,minimum height=1.5pt]

            \node[circle,draw,fill=lightgray, inner sep=0pt, minimum size=1.35cm, opacity=.3] (b) at (0.7,-0.000){};
            \node[circle,draw,fill=lightgray, inner sep=0pt, minimum size=1.35cm, opacity=.3] (c) at (1.50,-0.50){};
            \node[circle,draw,fill=lightgray, inner sep=0pt, minimum size=1.35cm, opacity=.3] (d) at (2.50, 0.25){};
            \node[circle,draw,fill=lightgray, inner sep=0pt, minimum size=1.35cm, opacity=.3] (e) at (3.00,-1.00){};
            \node[circle,draw,fill=lightgray, inner sep=0pt, minimum size=1.35cm, opacity=.3] (f) at (3.50,-0.75){};
            
            \node (b)[point] at (0.70,-0.000) {};
            \node (c)[point] at (1.50,-0.50) {};
            \node (e)[point] at (2.50, 0.25) {};
            \node (f)[point] at (3.00,-1.0) {};
            \node (g)[point] at (3.50,-0.75) {};

            \draw (g.center) -- (f.center);
            \draw (b.center) -- (c.center);
            \draw (c.center) -- (d.center);
            \draw (d.center) -- (f.center);
        \end{tikzpicture}}}}}\hfill
        \subfigure[Persistence Diagram\label{fig:vr:PD}]{\includegraphics[width=0.22\textwidth,height=!]{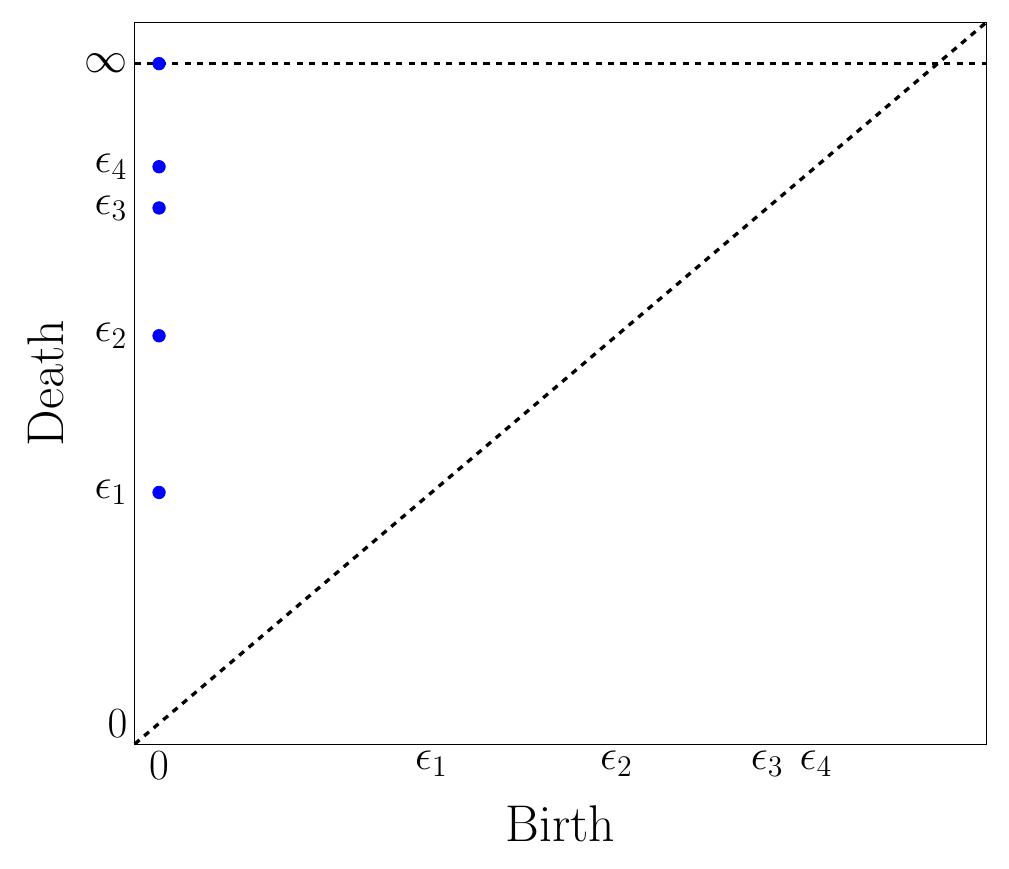}}\hfill
     \caption{\label{fig:vr} An example of computing the VR complex for a set of points in the Euclidean space. As the threshold increases, the $0$-simplices gradually vanish. In the end, we would have only one connected component that lives forever.}
     \vspace{0.10in}
\end{figure*}

\textit{Vietoris-Rips}~(VR)~\citep{vietoris1927hoheren} complex is a common method used for building a filtration of data points $\{\boldsymbol{a}_1, \boldsymbol{a}_2, \dots, \boldsymbol{a}_m\}$.
To construct such a filtration, we start by defining a metric space over which we compare our data points using some notion of distance $\mathrm{d}(\cdot, \cdot)$.
At each scale $\epsilon$, we connect all the data points that are within the $\epsilon$ distance of each other, i.e., data points that satisfy ${\mathrm{d}(\boldsymbol{a}_i, \boldsymbol{a}_j)\leq\epsilon}$.
We then sweep over all the possible $\epsilon$'s from $-\infty$ to $+\infty$ and record the birth and death of topological features of interest in a PD.
As mentioned earlier, in this paper we are mostly interested in capturing the 0D and 1D topological features.
Thus, we would have a separate PD for each dimension.
An illustrated example of finding the VR filtration for a set of point clouds is given in~\Cref{fig:vr}.

\section{Framework}
\label{sec:framework}
This section presents our framework for the topological analysis of shortcut learning in neural networks.
First, we formally introduce our problem setting and notation.
Then, we go over the details of the topological analysis of neural networks for shortcut learning.

\subsection{Problem Setting}
\label{sec:framework:setting}
Consider a labelled dataset ${\mathcal{D}=\{(\boldsymbol{x}_i, y_i)\}_{i=1}^{n}}$ where ${\boldsymbol{x}_i \in \mathbb{R}^{d}}$ and ${y_i}$ denote a data sample and its associated label.
Without loss of generality, in this paper, we consider a multi-class classification problem where $y_i$ belongs to exactly one class from the set $\{1, 2, \dots, k\}$.
Also, let $f_{\boldsymbol{\theta}}: \mathbb{R}^{d} \rightarrow \mathbb{R}^{k}$ denote a neural network classifier with parameters $\boldsymbol{\theta}$ that maps the inputs $\boldsymbol{x}$ to a real-valued vector $\boldsymbol{z} \in \mathbb{R}^{k}$, commonly known as the \textit{logit} vector.
The final prediction of the classifier is obtained via solving ${\hat{y} = \argmax_{c} \boldsymbol{z}[c]}$.
To train the classifier $f_{\boldsymbol{\theta}}$, we usually optimize a relevant objective over the dataset:
\begin{equation}\label{eq:train_obj}
    \boldsymbol{\theta}^{*} = \argmin_{\boldsymbol{\theta}} \sum_{i=1}^{n} \mathcal{L}\left(f_{\boldsymbol{\theta}}\left(\boldsymbol{x}_{i}\right), y_{i}\right),
\end{equation}
where $\mathcal{L}(\cdot, \cdot)$, called the loss function, is a measure of discrepancy between the prediction of the classifier and the ground-truth label.
In this paper, we assume that our loss function is a cross-entropy function.

To pave the way for our discussion of topological analysis of DNNs, we need to define a few extra notations.
Let $m$ be the total number of neurons in a DNN.
During inference, each input $\boldsymbol{x}_i$ is sent through the neural network and processed by its neurons.
Let $a_{j}^{(i)}$ denote the value of the $j$th neuron for the $i$th data sample $\boldsymbol{x}_i$.
We gather the values of the $j$th neuron for $N$ data samples in a so-called activation vector, written as:
\begin{equation}\label{eq:activation}
    \boldsymbol{a}_{j} = [a_{j}^{(1)}, a_{j}^{(2)}, \dots, a_{j}^{(N)}]^{\intercal}.
\end{equation}
Using this notation, if we get the activation values of all the $m$ neurons of the DNN for $N$ data samples, we end up having a collection of $m$ activation vectors ${\mathcal{A} = \{\boldsymbol{a}_1, \boldsymbol{a}_2, \dots, \boldsymbol{a}_m\}}$.
Next, we will see how we can create a topological map of a neural network computation graph using the set $\mathcal{A}$.

\subsection{Topological Analysis of Shortcut Learning}
\label{sec:framework:TDA}
As mentioned before, shortcut learning is commonplace in neural networks.
Intuitively, this phenomenon happens when a DNN learns an unusually direct path from its input to its output.
We aim to reveal this unusual behavior by analyzing the traversal of input information through the computational graph of the neural network.
We resort to PH for identifying this abnormal behavior in DNNs.
To this end, we construct a VR filtration of the neural network computation graph following a method introduced in~\citet{zheng2021topobackdoor}.

Recall from \Cref{sec:background:vietoris} that the first step during the construction of a VR filtration is to determine a meaningful distance measure between the data points.
In our analysis, we are dealing with neurons as nodes in a computational graph.
A shortcut happens when there are \textit{multiple} neurons throughout the graph that are activated together.
For this reason, we can use a standard correlation measure between the activation vectors of two different neurons as our distance metric in building a VR complex.

In other words, let us assume that $\boldsymbol{a}_{i}$ and $\boldsymbol{a}_{j}$ are two different activation vectors ($i \neq j$) constructed according to \Cref{eq:activation}.
In addition, let $-1 \leq \rho(\boldsymbol{a}_{i}, \boldsymbol{a}_{j}) \leq 1$ be a normalized correlation metric between these two vectors.
Then, we define the distance between these two activation vectors as:
\begin{equation}\label{eq:vr_distance}
    \mathrm{d}(\boldsymbol{a}_i, \boldsymbol{a}_j) = 1 - \rho(\boldsymbol{a}_{i}, \boldsymbol{a}_{j}).
\end{equation}
This means that if the activation vectors $\boldsymbol{a}_{i}$ and $\boldsymbol{a}_{j}$ are highly correlated, their distance is smaller and vice versa.

Using this distance metric, we next build a VR filtration for the set of all neurons $\mathcal{A}$.
To this end, we assume that each neuron is denoted by a node in a graph.
Then, we sweep over $\epsilon$ from $-\infty$ to $+\infty$ and connect all the neurons whose activation vectors fall within an $\epsilon$ distance of each other.
This process would result in a filtration $\emptyset \subseteq \mathcal{G}_{\epsilon_1} \subseteq \mathcal{G}_{\epsilon_2} \subseteq \dots \subseteq \mathcal{G}_{\epsilon_{\infty}}$.
Here, $\mathcal{G}_{\epsilon_i}$ is a graph representation of neurons, where the edges are connected to each other if they have a distance less than $\epsilon_i$. 
We then use existing TDA tools to extract the 0D and 1D persistence diagrams of this filtration.

We are particularly interested in 1D topological structures, or cycles, in the computational graphs.
This is because such structures capture a subset of neurons that are highly correlated with each other, meaning that they are often activated together.
Remember that our activation vectors have been computed for $N$ different inputs, and if $N$ is large enough, such activation vectors should give us a fairly comprehensive overview of input traversal through neurons in a DNN.
As such, finding a subset of neurons that are activated together (which could create a cycle in the topological features of computational graphs) can signal a shortcut path.


\section{Case Studies}
\label{sec:cases}

Now that we have established a framework for extracting the topological features of DNNs, in this section we show how the same framework is applicable to two different issues that both arise as a result of shortcut learning.
We also present an extra case study on backdoor attacks in the Appendix.

\subsection{Case Study I: Unlearnable Examples}
\label{sec:cases:unlearnable}

\subsubsection{Overview}
Unlearnable examples~\citep{huang2021emn}, also known as availability attacks, are a family of data poisoning attacks~\citep{biggio2018wild} aiming to protect personal data from being exploited.
With the unauthorized access of third-party crawlers over the web, users might lose control over how their personal data is used for training deep neural networks such as automated facial recognition~\citep{hill2019photos, hill2020secretive}. 
The goal of unlearnable examples is to passively prevent this misuse by adding an imperceptible perturbation to the user's data before sharing it.
This perturbation should be powerful enough not to let any DNN learn meaningful patterns from its underlying data, but imperceptible enough so it does not interfere with the utility of the data to be shared with its audience~\citep{huang2021emn}.

There are various ways to construct such data-protecting perturbations~\citep{huang2021emn, yuan2021ntga, fowl2021tap, fu2022remn, yu2022shr, sandoval2022ar}.
In general, the goal of all these different approaches is to find a set of perturbations $\boldsymbol{\delta}_{i}$ for each training data point $\boldsymbol{x}_{i}$ such that after optimizing the neural network weights over the perturbed data, the trained network yields a high error rate over the test set.\footnote{Note that the data protector might opt to protect only a subset of all the training data. In this case, we can simply set $\boldsymbol{\delta}_{i} = \boldsymbol{0}$ for data samples that are not going to be protected.}
We can write this objective using our notations as:
\begin{align}\label{eq:threat_model}
    \argmax_{\{\boldsymbol{\delta}_{i}~|~i \in \mathrm{train}\}} &\sum_{j \in \mathrm{test}} \mathcal{L}\left(f_{\boldsymbol{\theta}^{*}}(\boldsymbol{x}_{j}), y_{j}\right) \\\nonumber
    &\text{s.t.}~\boldsymbol{\theta}^{*} = \argmin_{\boldsymbol{\theta}} \sum_{i \in \mathrm{train}} \mathcal{L}\left(f_{\boldsymbol{\theta}}\left(\boldsymbol{x}_{i} + \boldsymbol{\delta}_{i}\right), y_{i}\right).
\end{align}
Different methods opt to achieve this goal from different perspectives.
One approach that is closely related to our analysis is presented in~\citet{yu2022shr}.
It is argued that the majority of existing unlearnable examples create shortcut paths within the neural network during training~\citep{yu2022shr,segura2023what}.
As such, the perturbation generation process can be oversimplified, and even a set of linearly separable noisy data can act as unlearnable perturbations~\citep{yu2022shr}.
Therefore, since unlearnable examples are related to shortcut learning in neural networks, they can be an interesting case study for our analysis.

\subsubsection{Experimental Evaluation}

\begin{figure*}[tb!]
        \centering
        \subfigure[TAP]{\includegraphics[width=0.225\textwidth,height=!]{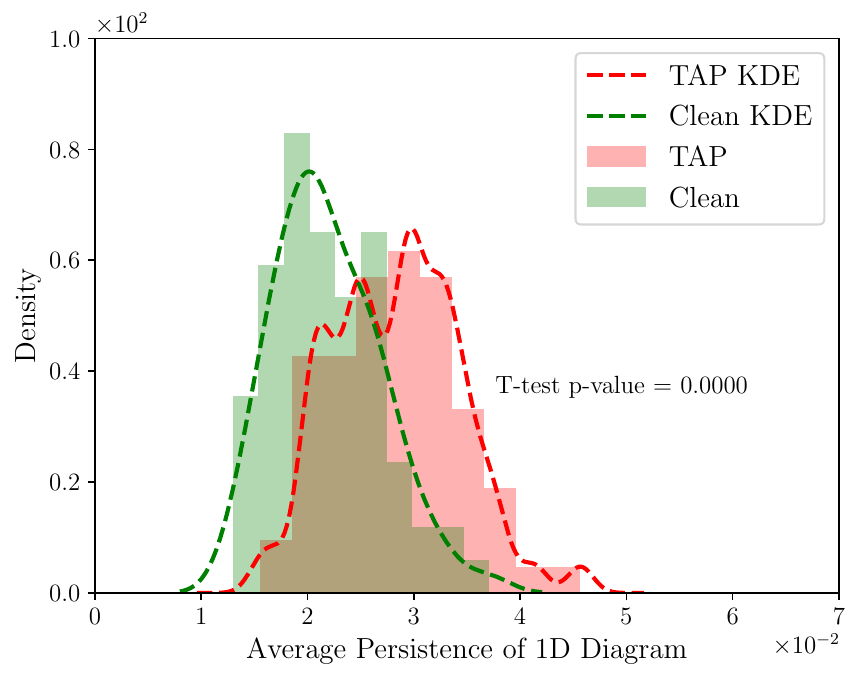}}\hspace{1.75em}
        \subfigure[SHR]{\includegraphics[width=0.225\textwidth,height=!]{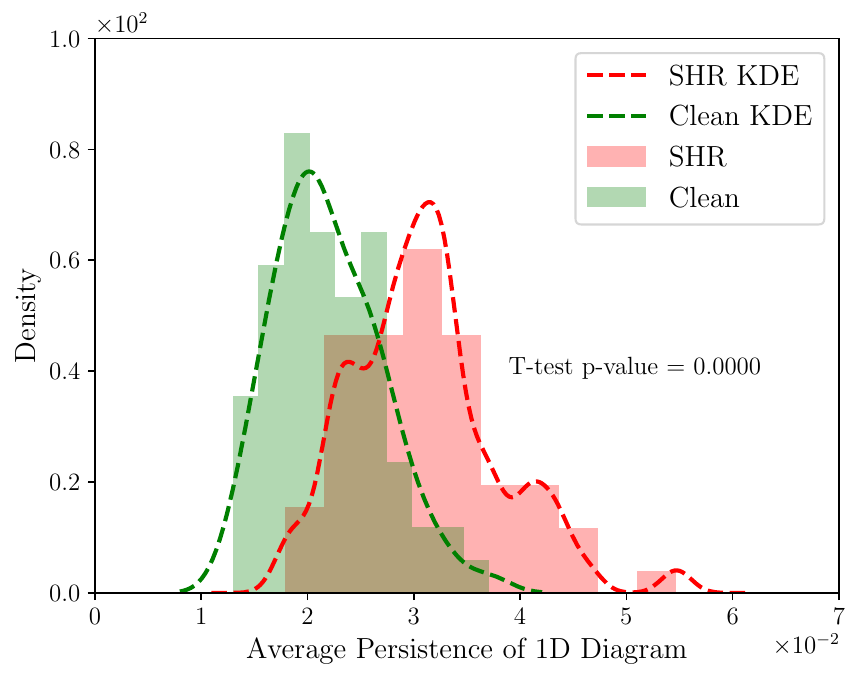}}\hspace{1.75em}
        \subfigure[REM]{\includegraphics[width=0.225\textwidth,height=!]{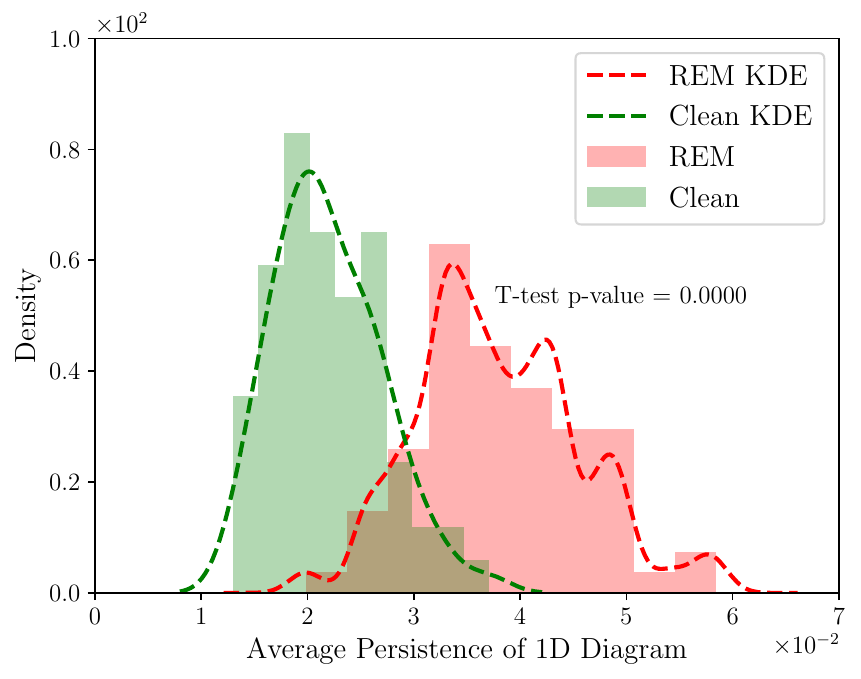}}\\\vspace{-1em}
        \subfigure[EMN]{\includegraphics[width=0.225\textwidth,height=!]{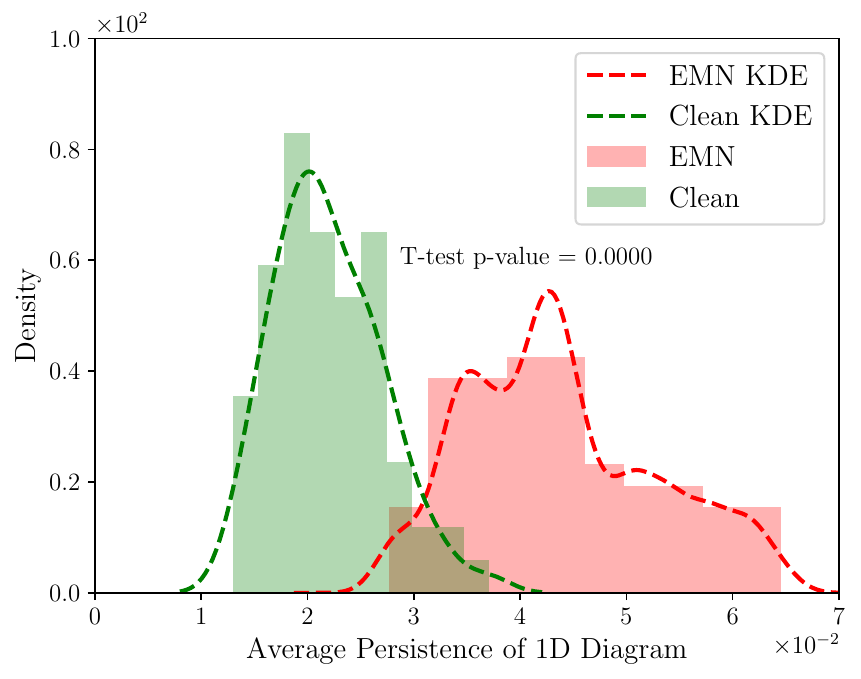}}\hspace{1.75em}
        \subfigure[NTGA]{\includegraphics[width=0.225\textwidth,height=!]{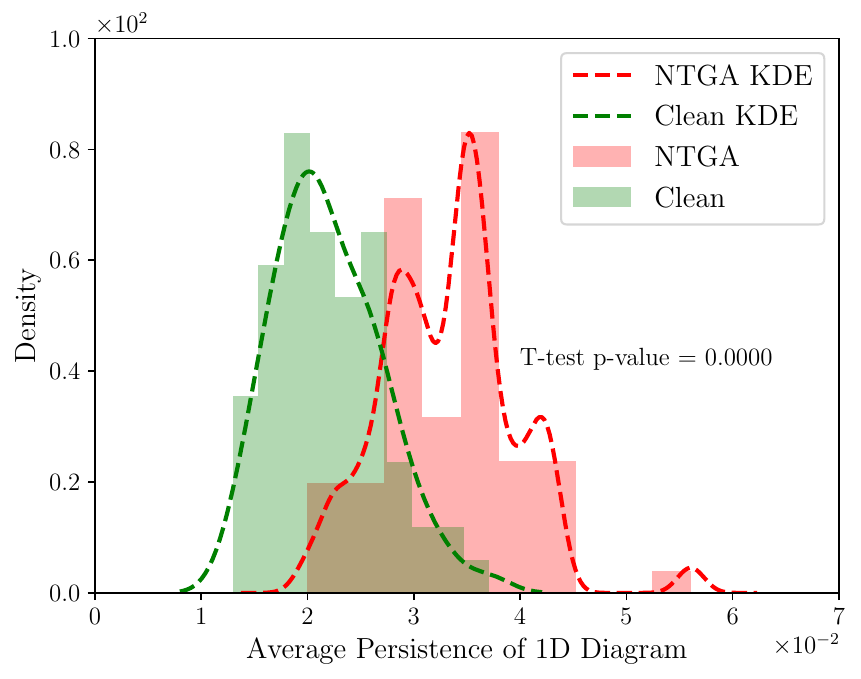}}\hspace{1.75em}
        \subfigure[AR]{\includegraphics[width=0.225\textwidth,height=!]{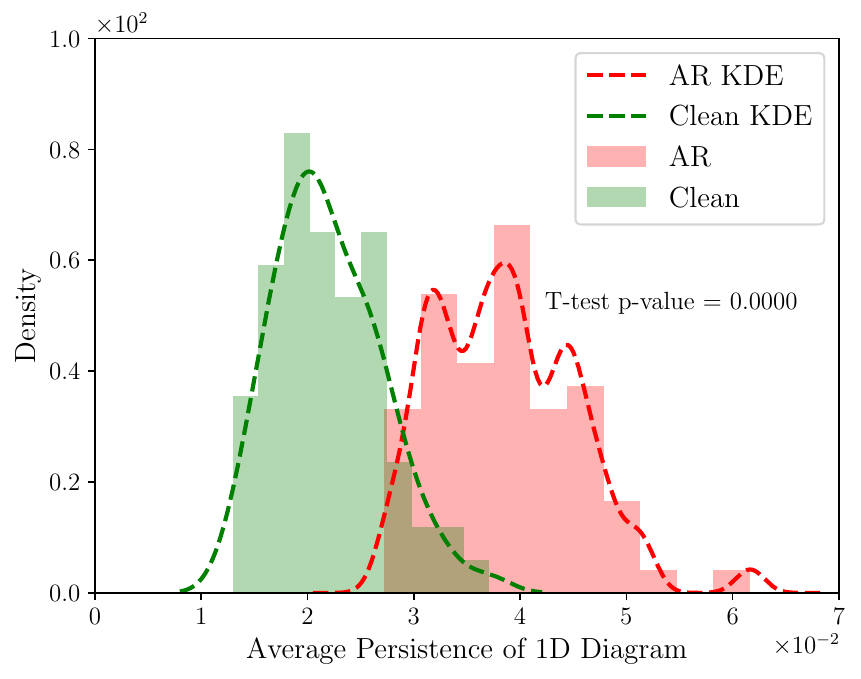}}
     \caption{\label{fig:ttest} The distribution of average persistence of 1D homology groups for ResNet-18 models trained with different versions of unlearnable CIFAR-10 datasets. Each histogram summarizes the result of 70 independent training runs for each dataset. The p-value of the T-test has also been shown in each figure.}
\end{figure*}

\paragraph{Settings:}
To empirically validate our speculations for the existence of a topological signature within DNNs trained over unlearnable datasets, we choose six state-of-the-art methods used for generating unlearnable perturbations.
These approaches include Targeted Adversarial Poisoning~(TAP)~\citep{fowl2021tap}, Error-minimizing Noise~(EMN)~\citep{huang2021emn}, Robust EMN~(REM)~\citep{fu2022remn}, Neural Tangent Generalization Attacks~(NTGA)~\citep{yuan2021ntga}, Shortcut~(SHR)~\citep{geirhos2020shortcut}, and Autoregressive attacks~(AR)~\citep{sandoval2022ar}.
Using these methods, we create unlearnable CIFAR-10~\citep{krizhevsky2009learning} datasets and train 70 ResNet-18~\citep{he2016deep} models over each dataset independently.
Furthermore, we also train 70 ResNet-18 models over the clean version of the CIFAR-10 dataset.
We use the settings used in~\citet{huang2021emn} for training our models.

\paragraph{Evaluation:}
After training each model until convergence, we use our framework outlined in~\Cref{sec:framework} to extract topological features of the DNN computational graph.
We use the \texttt{ripser} package~\citep{ctralie2018ripser} to obtain the 1D persistence diagram~(PD) of the computational graph using a VR filtration.
Then, we operate over this PD for a statistical analysis of the topological features.
To this end, we use the average life-span/persistence of the PD~(\textsc{Avg PD\textsubscript{1}}) as well as the Wasserstein distance~(\textsc{WSD}) between a pair of PDs as our evaluation metrics~\citep{gudhi2015}.

\begin{figure*}[tb!]
        \centering
        \rotatebox[x=1em,y=2.5em]{90}{Rank 1\label{fig:trajectories_appendix:rank1}}
        \subfigure{\includegraphics[width=0.13\textwidth,height=!]{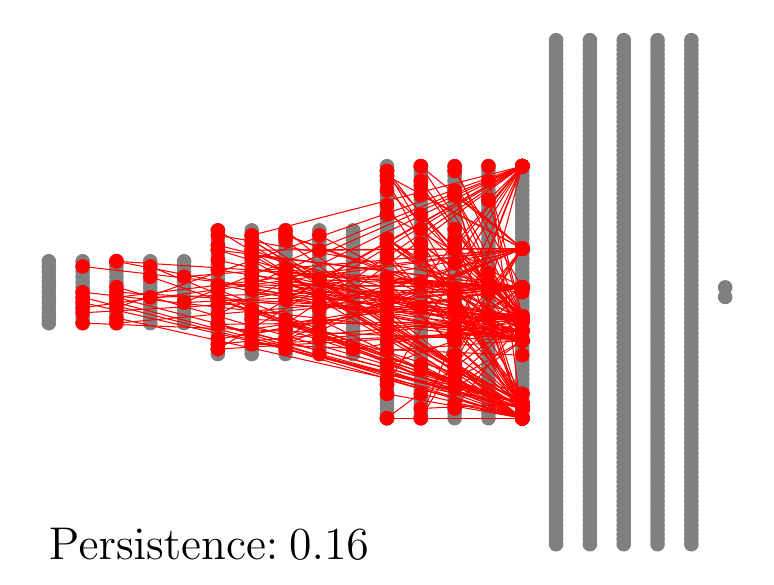}}\hspace{0.1em}
        \subfigure{\includegraphics[width=0.13\textwidth,height=!]{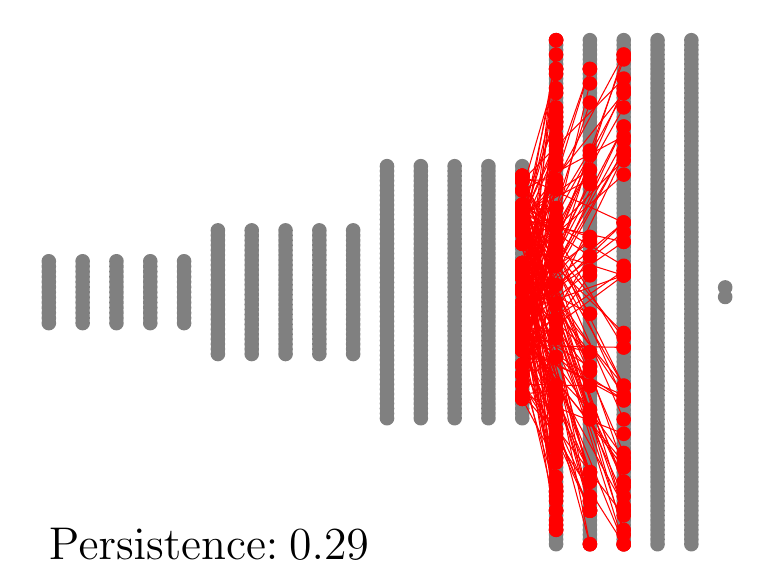}}\hspace{0.1em}
        \subfigure{\includegraphics[width=0.13\textwidth,height=!]{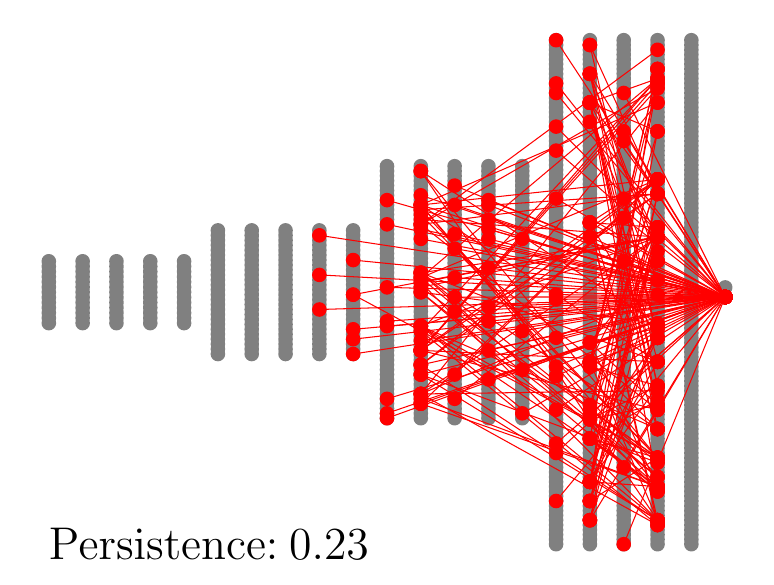}}\hspace{0.1em}
        \subfigure{\includegraphics[width=0.13\textwidth,height=!]{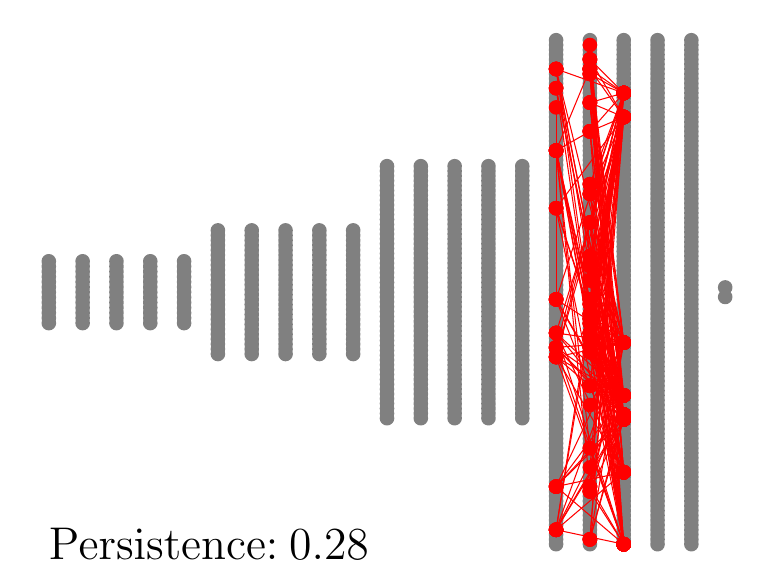}}\hspace{0.1em}
        \subfigure{\includegraphics[width=0.13\textwidth,height=!]{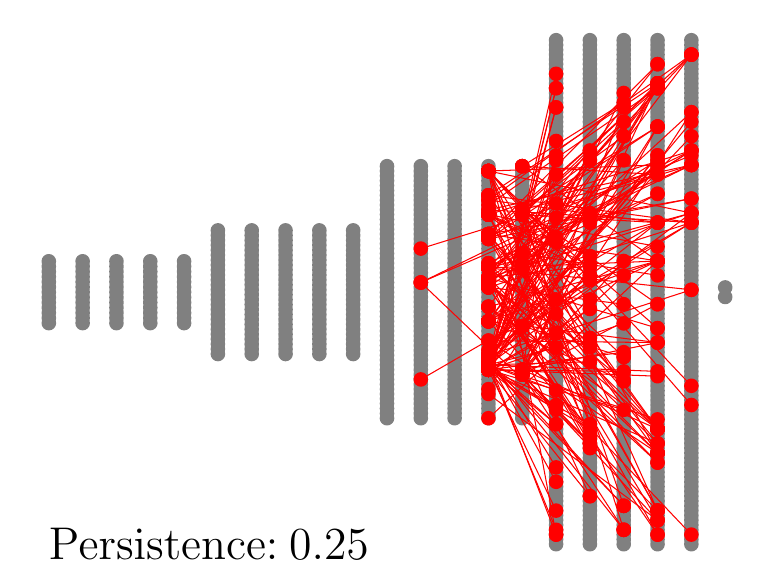}}\hspace{0.1em}
        \subfigure{\includegraphics[width=0.13\textwidth,height=!]{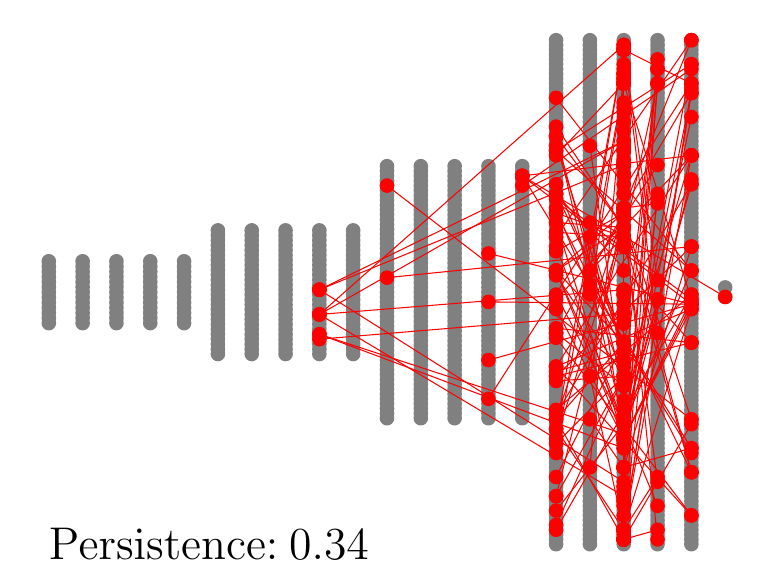}}\hspace{0.1em}
        \subfigure{\includegraphics[width=0.13\textwidth,height=!]{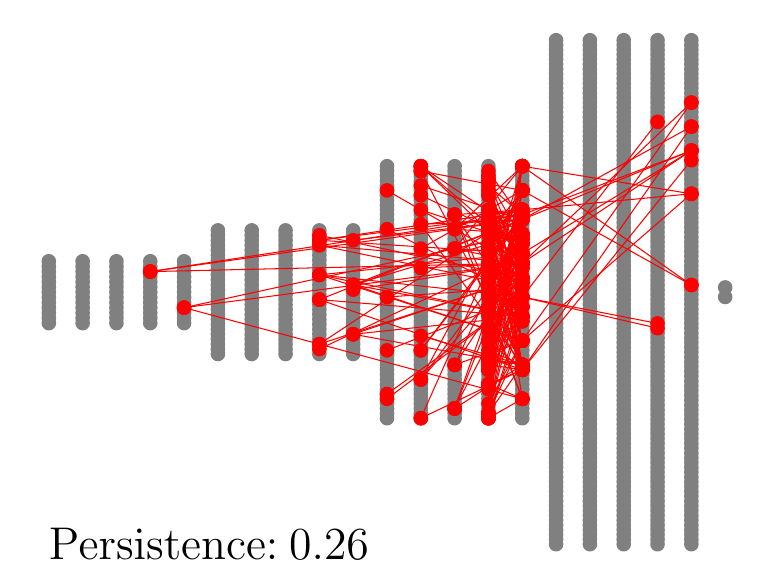}}\\
        \rotatebox[x=1em,y=2.5em]{90}{Rank 2\label{fig:trajectories_appendix:rank2}}
        \subfigure{\includegraphics[width=0.13\textwidth,height=!]{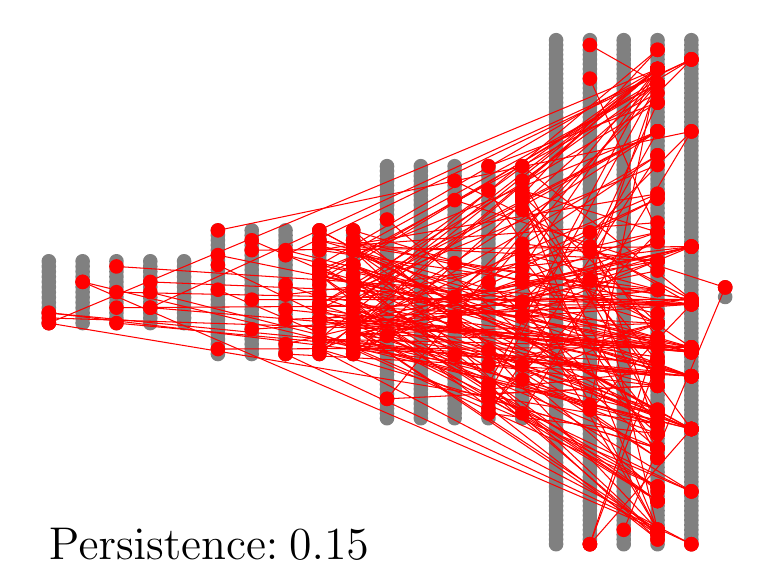}}\hspace{0.1em}
        \subfigure{\includegraphics[width=0.13\textwidth,height=!]{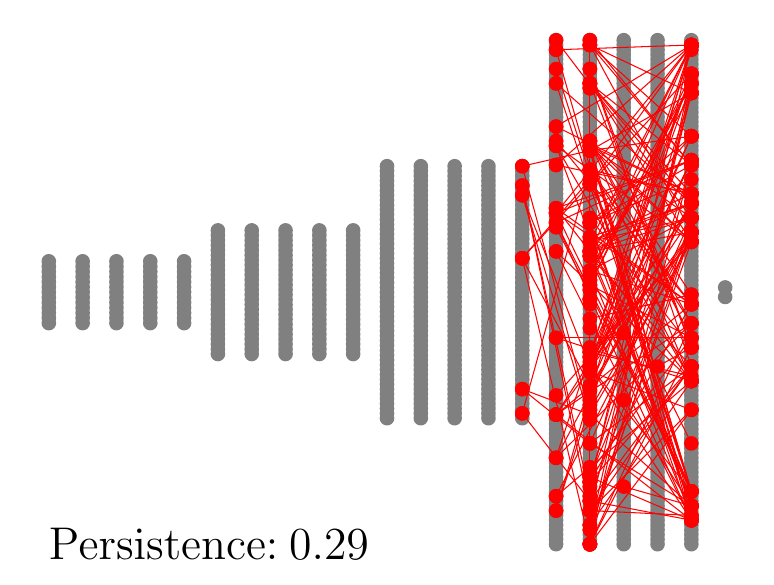}}\hspace{0.1em}
        \subfigure{\includegraphics[width=0.13\textwidth,height=!]{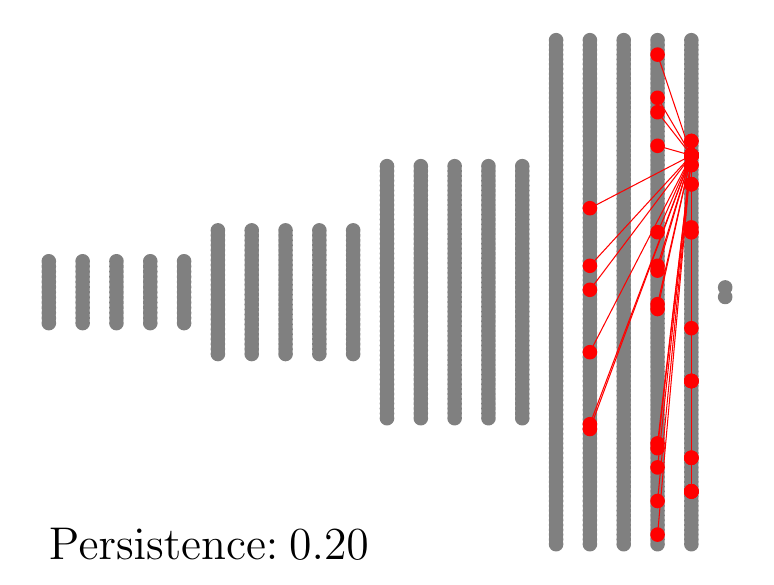}}\hspace{0.1em}
        \subfigure{\includegraphics[width=0.13\textwidth,height=!]{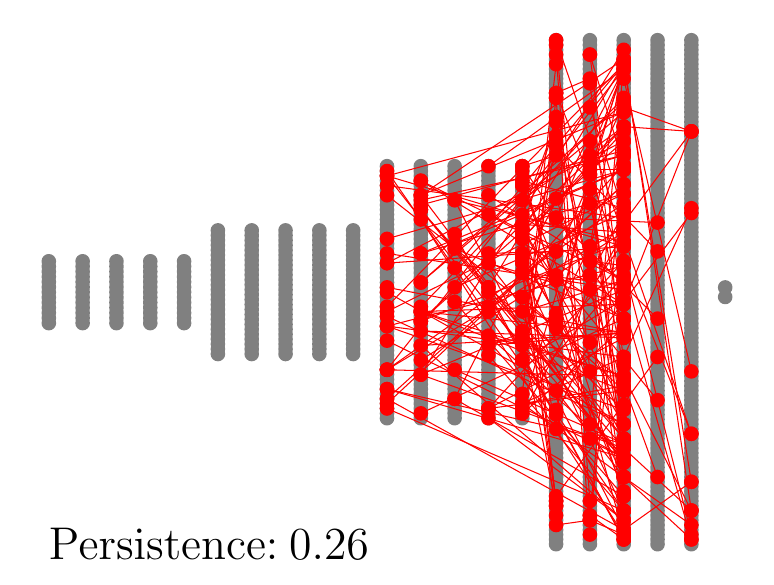}}\hspace{0.1em}        
        \subfigure{\includegraphics[width=0.13\textwidth,height=!]{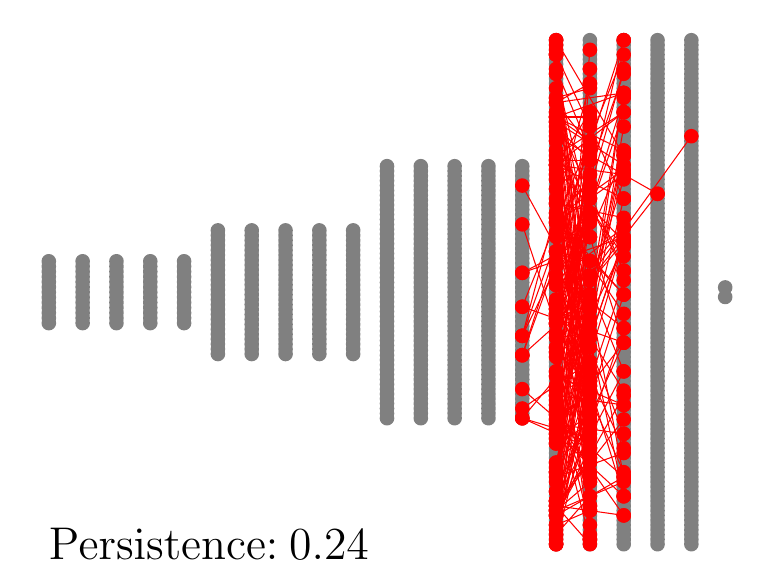}}\hspace{0.1em}
        \subfigure{\includegraphics[width=0.13\textwidth,height=!]{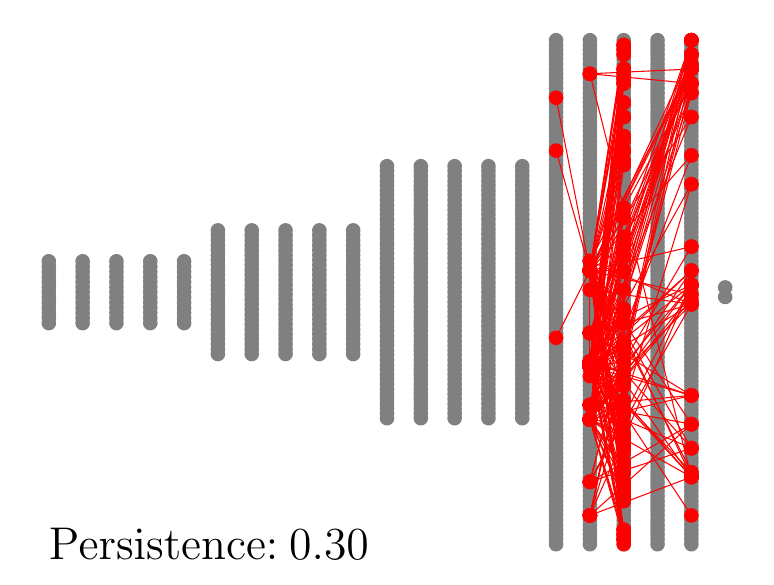}}\hspace{0.1em}
        \subfigure{\includegraphics[width=0.13\textwidth,height=!]{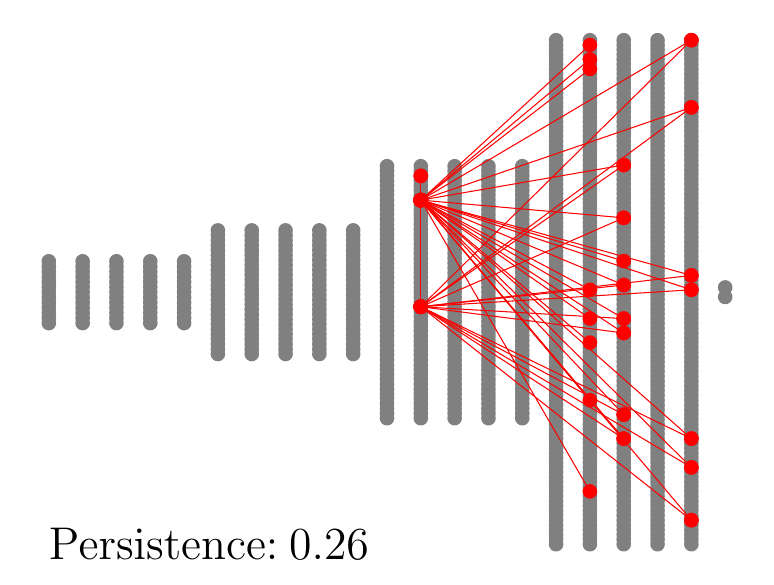}}\\
        \rotatebox[x=1em,y=2.5em]{90}{Rank 3\label{fig:trajectories_appendix:rank3}}
        \setcounter{subfigure}{0}
        \subfigure[Clean]{\includegraphics[width=0.13\textwidth,height=!]{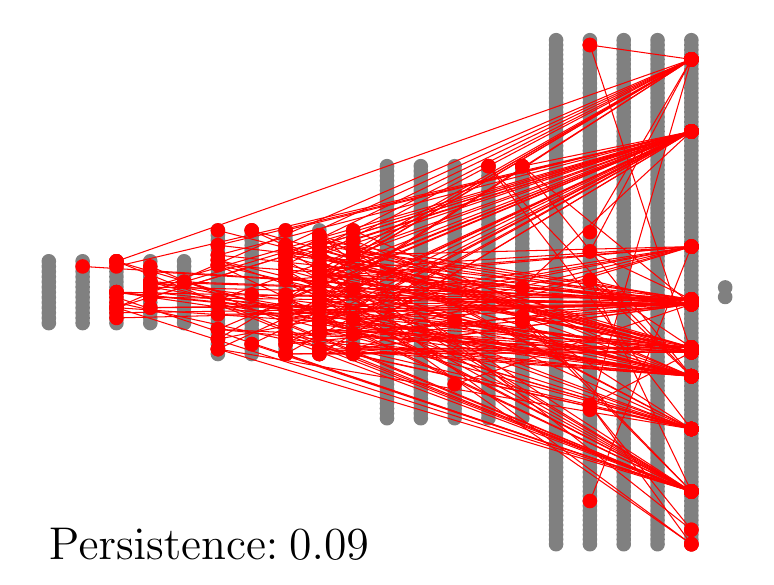}}\hspace{0.1em}
        \subfigure[TAP]  {\includegraphics[width=0.13\textwidth,height=!]{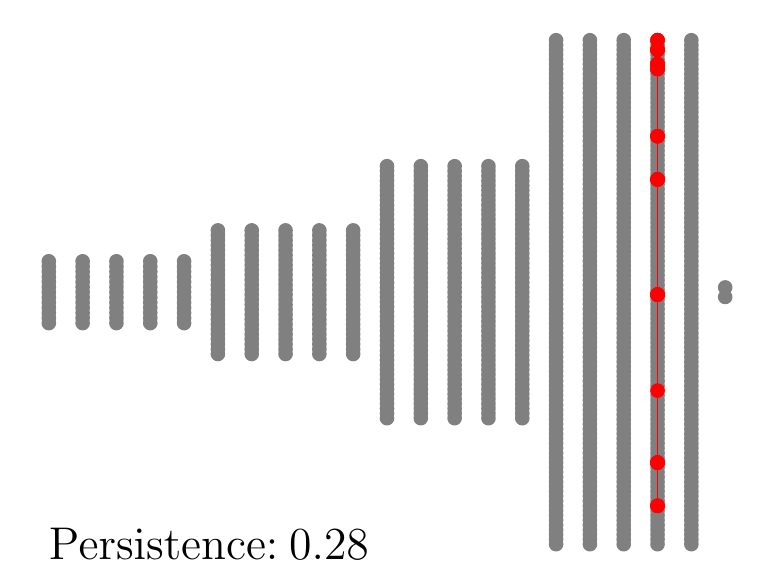}}\hspace{0.1em}
        \subfigure[SHR]  {\includegraphics[width=0.13\textwidth,height=!]{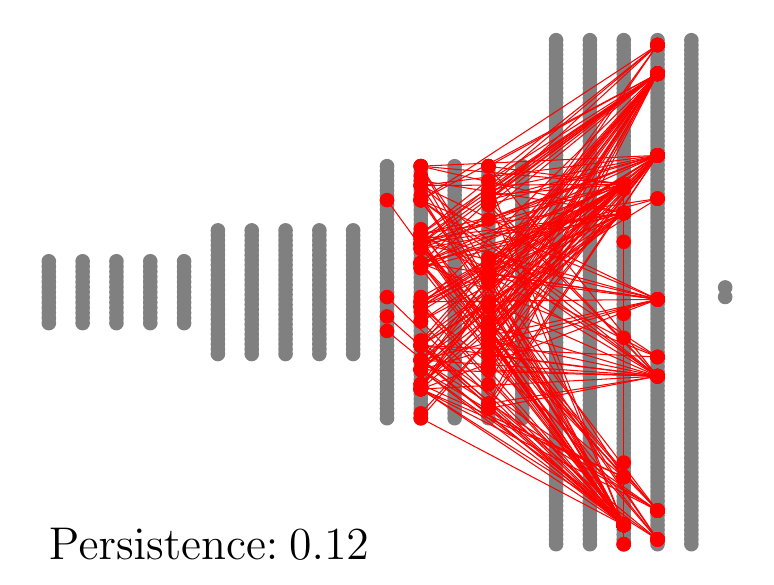}}\hspace{0.1em}
        \subfigure[REM]  {\includegraphics[width=0.13\textwidth,height=!]{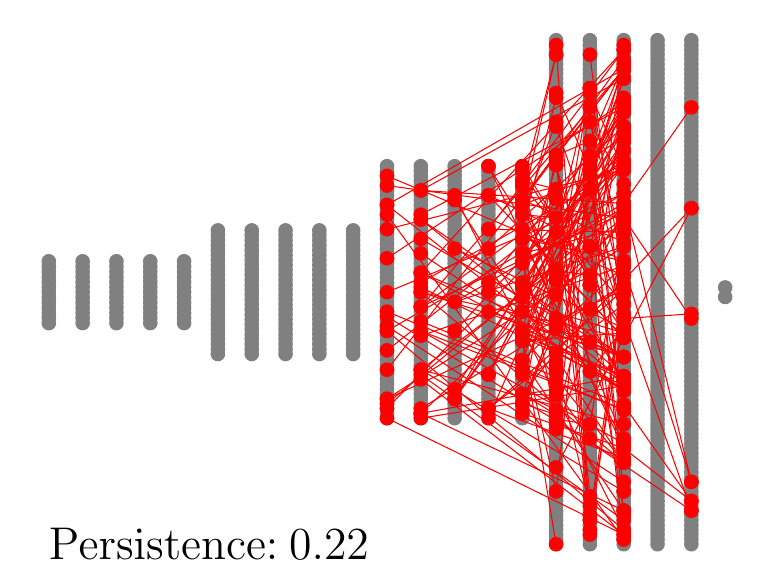}}\hspace{0.1em}
        \subfigure[EMN]  {\includegraphics[width=0.13\textwidth,height=!]{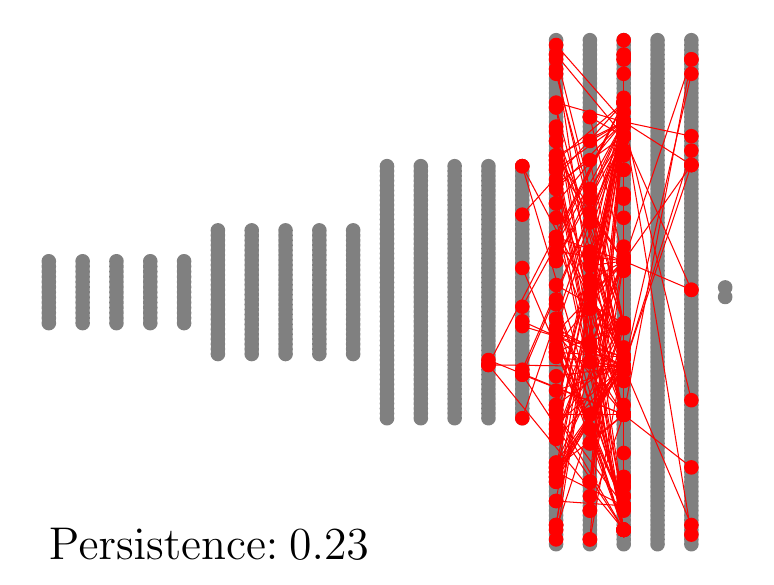}}\hspace{0.1em}
        \subfigure[NTGA] {\includegraphics[width=0.13\textwidth,height=!]{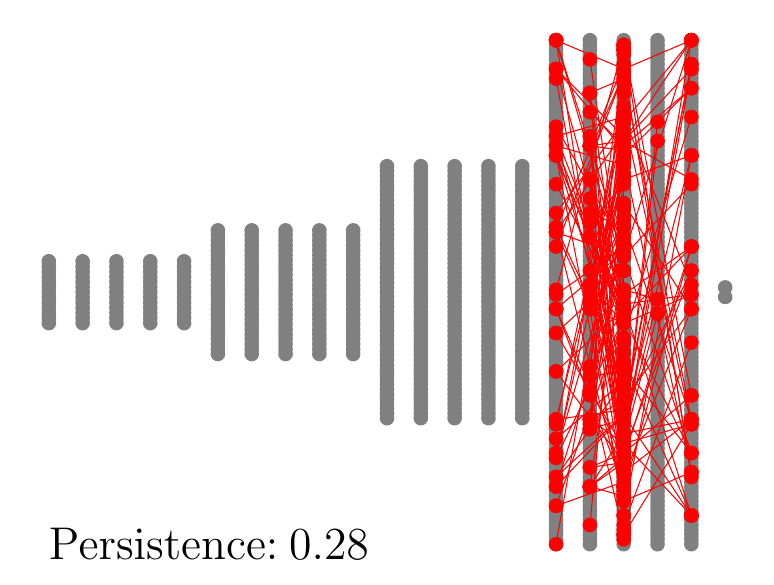}}\hspace{0.1em}
        \subfigure[AR]   {\includegraphics[width=0.13\textwidth,height=!]{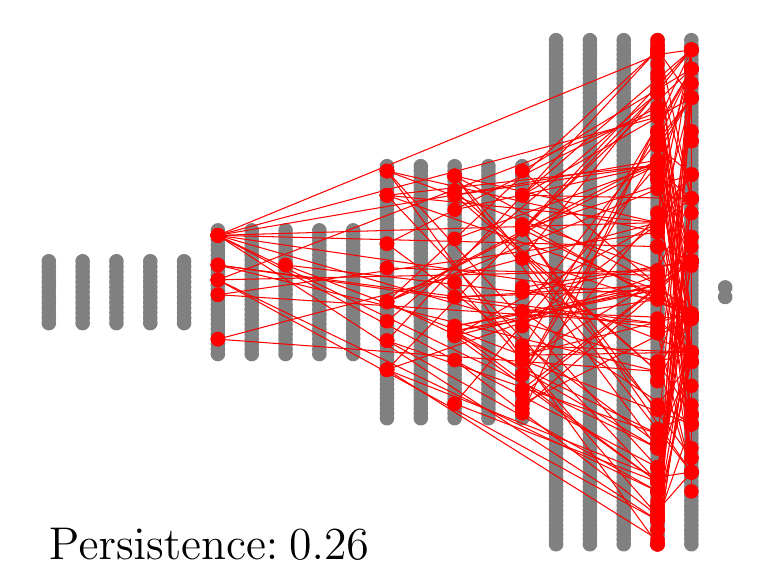}}
     \caption{\label{fig:trajectories} Top-3 persistent cycles for one instance of ResNet-18 models trained with different versions of unlearnable CIFAR-10 datasets. Each node denotes a single neuron within the model. Note that due to the high number of neurons, we show a downsampled version of the original model.}
     \vspace{0.2in}
\end{figure*}

\begin{figure*}[tb!]
        \centering
        \rotatebox[x=1em,y=2.5em]{90}{(a)\label{fig:features_appendix:clean}}
        \subfigure{\includegraphics[width=0.96\textwidth,height=!]{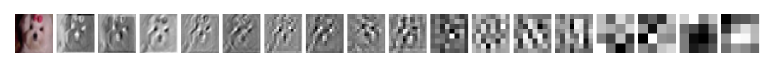}}\\\vspace{-2em}
        \rotatebox[x=1em,y=2.5em]{90}{(b)\label{fig:features_appendix:tap}}
        \subfigure{\includegraphics[width=0.96\textwidth,height=!]{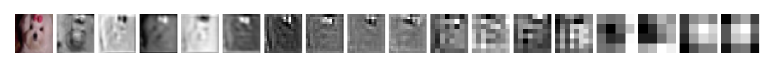}}\\\vspace{-2em}
        \rotatebox[x=1em,y=2.5em]{90}{(c)\label{fig:features_appendix:shr}}
        \subfigure{\includegraphics[width=0.96\textwidth,height=!]{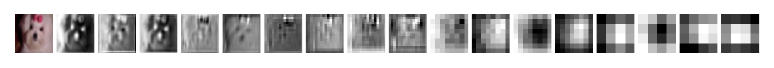}}\\\vspace{-2em}
        \rotatebox[x=1em,y=2.5em]{90}{(d)\label{fig:features_appendix:rem}}
        \subfigure{\includegraphics[width=0.96\textwidth,height=!]{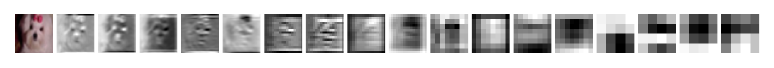}}\\\vspace{-2em}
        \rotatebox[x=1em,y=2.5em]{90}{(e)\label{fig:features_appendix:emn}}
        \subfigure{\includegraphics[width=0.96\textwidth,height=!]{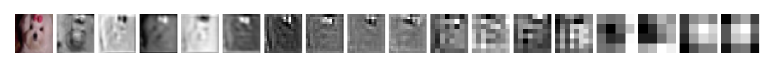}}\\\vspace{-2em}
        \rotatebox[x=1em,y=2.5em]{90}{(f)\label{fig:features_appendix:ntga}}
        \subfigure{\includegraphics[width=0.96\textwidth,height=!]{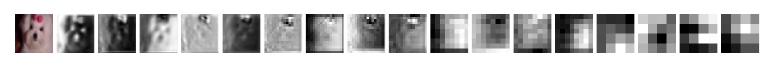}}\\\vspace{-2em}
        \rotatebox[x=1em,y=2.5em]{90}{(g)\label{fig:features_appendix:ar}}
        \subfigure{\includegraphics[width=0.96\textwidth,height=!]{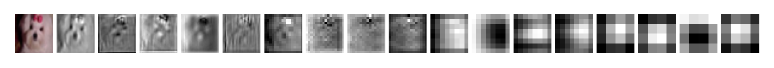}}
     \caption{\label{fig:features} Feature space representation of one instance of ResNet-18 models trained with different versions of unlearnable CIFAR-10 datasets. The features are shown from the input space~(leftmost) to the last layer of the classifier~(rightmost). The models are trained using (a)~clean, (b)~TAP~\citep{fowl2021tap}, (c)~SHR~\citep{yu2022shr}, (d)~REM~\citep{fu2022remn} (e)~EMN~\citep{huang2021emn}, (f)~NTGA~\citep{yuan2021ntga}, and (g)~AR~\citep{sandoval2022ar} data.}
\end{figure*}

\begin{figure*}[tb!]
\centering
    \subfigure[ResNet-18]{\includegraphics[width=0.3\textwidth]{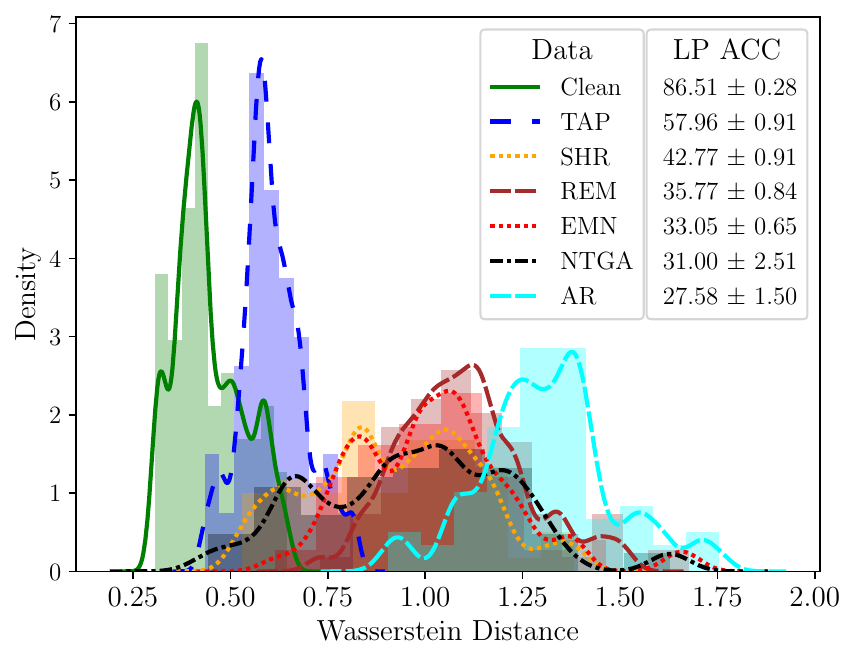}}\hspace{2em}
    \subfigure[DenseNet-121]{\includegraphics[width=0.3\textwidth]{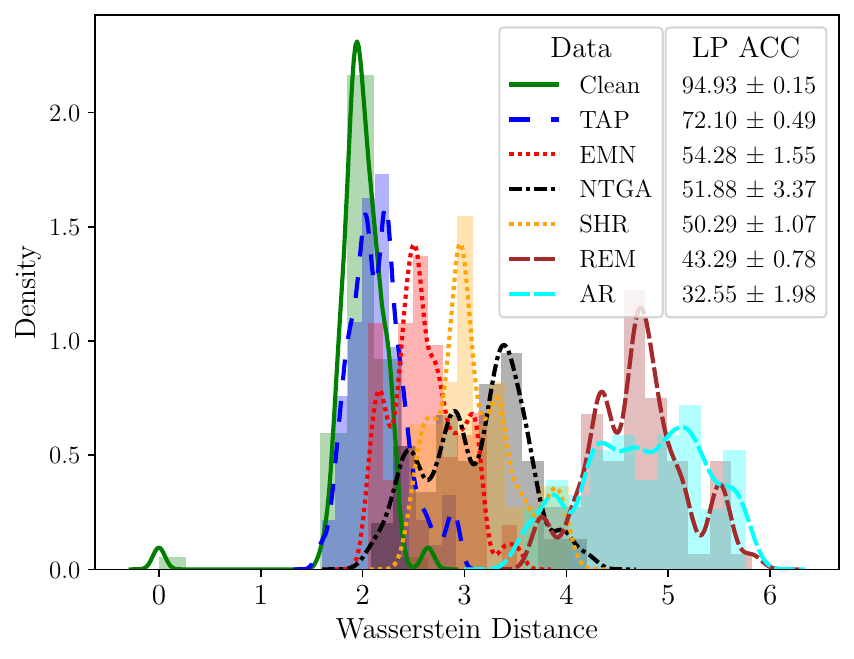}}
	\caption{\label{fig:wasserstein} The Wasserstein distance between 1D persistence diagrams of clean versus unlearnable (a) ResNet-18 and (b) DenseNet-121 models. For clean models, we compute the distance between two randomly selected models, while for unlearnable models we use a clean baseline.}
\end{figure*}

\paragraph{Findings:}
We can summarize our key findings as below:
\vspace{-0.5em}
\begin{itemize}\setlength\itemsep{0.2em}
    \item As shown in~\Cref{fig:ttest}, models trained on unlearnable datasets exhibit a different average persistence compared to clean models.
    The difference is statistically significant for all the tested unlearnable models, where the reported p-value is almost zero.
    This result demonstrates a clear distinguishing factor between benign and unlearnable models in their topological features, showing that \textbf{unlearnable datasets create alternative trajectories within the DNN for processing the inputs}.
    These paths have a higher life-span which indicates the over-confidence of the model in utilizing them. 
    \item Glancing at the highly persistent cycles within each model as shown in~\Cref{fig:trajectories}, we can see that \textbf{unlearnable models block the flow of information from the clean test inputs}.
    Therefore, the models trained over unlearnable datasets should indeed exhibit a lower accuracy compared to the clean model.
    This observation can be verified by looking at the feature maps of the ResNet-18 models for a sample input.
    As shown in~\Cref{fig:features}, for unlearnable models we do not see any trace of the input during the intermediate layers, while this information is easily transferred to deeper layers of the clean DNN.
    \item \citet{segura2023what} argued that even unlearnable models capture useful features that can be utilized for a better classification.
    To this end, they used a simple linear classifier to train over the frozen feature extractor of such models with a small validation set.
    \citet{kirichenko2023spurious} earlier showed that this approach is useful in combating spurious correlations.
    Using the same approach, we train a linear classifier head to obtain a linear probe accuracy~(\textsc{LP Acc}) for all our models.
    We also compute the Wasserstein distance between the PD of each model and a randomly sampled clean model.
    As can be seen in~\Cref{fig:wasserstein}, \textbf{there is a correspondence between the power of the unlearnable dataset and its ability to create alternative trajectories within the DNN}.
    In particular, a higher Wasserstein distance indicates that the trajectories of the trained model differ more from a baseline clean model.
    Therefore, models with a higher Wasserstein distance tend to have a lower LP accuracy, which shows that the underlying unlearnable dataset is more resilient.
    For an extended version of our results, please see the Appendix.
\end{itemize}

\subsection{Case Study II: Bias and Fairness}
\label{sec:cases:bias}

\subsubsection{Overview}
Another interesting case of shortcut learning happens during the training of biased neural networks~\citep{geirhos2020shortcut}.
It is widely known that bias usually takes place when a model focuses on one or a group of sensitive attributes to make its final prediction while ignoring the rest of input features~\citep{hardt2023fairness}.
Take the loan approval application as an example.
In this case, bias happens when the model puts extra focus on a certain sensitive attribute, like ZIP code, to decide whether a loan request should be approved or not.
Even though this notion is easy to comprehend, it is usually hard to track how the features are being processed within a machine learning model, especially when it comes to DNN models in the vision domain.

We can see traces of shortcut learning in the issue of bias and fairness in neural networks.
Going back to our example of loan approvals, we can see that the ZIP code can act as a spurious feature: there exists a shortcut within the neural network which is activated when certain sensitive ZIP codes appear as the input.
For these ZIP codes, the model ignores the rest of the input features and outputs a negative decision.
Even though this issue is quite different from the previous case study, we demonstrate that similar observations can be made through the lens of persistent homology.

\subsubsection{Experimental Evaluation}

\begin{table*}[tb!]
	\caption{Performance of ResNet-18 models trained over CelebA dataset with different target attributes. The sensitive attribute for all models is gender. In each case, we train a regular and an unbiased model using~\citep{seo2022unsupervised}. The results are averaged over 10 models.}
    \label{tab:fairness}
	\begin{center}
		\begin{footnotesize}
            \begin{sc}
		    \setlength\tabcolsep{0.5em}
			\def\arraystretch{1.0}
			\begin{tabular}{lccccccc}
				\toprule
                \multirow{3}{*}{\textbf{\shortstack[c]{Target\\Attribute}}}
                &\multirow{3}{*}{\textbf{Model}}
				&\multicolumn{6}{c}{\textbf{Performance Metrics}}\\
				\cmidrule(lr){3-8}
				&                                      & \shortstack[c]{Unbiased\\Acc~(\%)} & \shortstack[c]{Worst Group\\Acc~(\%)}  & \shortstack[c]{Unbiased\\Acc Std.~(\%)} & \shortstack[c]{EO\\Disparity~(\%)} & \shortstack[c]{Average\\Odds~(\%)}     & \shortstack[c]{Top-5~1D\\Pers.~($\times 10^{-1}$)} \\
				\midrule
                \multirow{2}{*}{\textbf{\shortstack[l]{Blonde\\Hair}}}    & Biased   & $84.50 \pm 1.35$  & $54.81 \pm 5.76$      & $17.39 \pm 2.56$       & $37.13 \pm 4.47$  & $21.31 \pm 2.23$  & $1.38 \pm 0.45$ \\
                                                                         & Unbiased & $84.08 \pm 2.26$  & $78.20 \pm 4.17$      & $5.84 \pm 2.31$        & $12.89 \pm 5.83$  & $8.83 \pm 4.39$   & $0.88 \pm 0.24$ \\
                \midrule
                \multirow{2}{*}{\textbf{Chubby}}                         & Biased   & $67.22 \pm 1.73$  & $22.24 \pm 1.64$      & $31.35 \pm 1.81$       & $32.13 \pm 7.28$  & $19.50 \pm 4.59$  & $1.06 \pm 0.41$ \\
                                                                         & Unbiased & $73.42 \pm 2.24$  & $61.53 \pm 5.01$      & $11.39 \pm 2.85$       & $19.53 \pm 6.88$  & $22.11 \pm 6.28$  & $0.70 \pm 0.22$ \\
                \midrule
                \multirow{2}{*}{\textbf{\shortstack[l]{Double\\Chin}}}   & Biased   & $66.76 \pm 2.96$  & $21.02 \pm 6.04$      & $32.04 \pm 3.55$       & $32.10 \pm 6.39$  & $19.19 \pm 4.06$  & $1.05 \pm 0.27$ \\
                                                                         & Unbiased & $75.81 \pm 1.49$  & $65.58 \pm 3.97$      & $8.86 \pm 2.76$        & $14.65 \pm 5.84$  & $17.16 \pm 5.66$  & $0.74 \pm 0.18$ \\
                \midrule
                \multirow{2}{*}{\textbf{\shortstack[l]{Heavy\\Makeup}}}  & Biased   & $74.58 \pm 1.36$  & $49.29 \pm 4.05$      & $22.05 \pm 1.66$       & $43.71 \pm 5.29$  & $43.59 \pm 3.39$  & $1.13 \pm 0.27$ \\
                                                                         & Unbiased & $74.69 \pm 1.93$  & $56.81 \pm 2.35$      & $15.62 \pm 1.92$       & $23.87 \pm 6.02$  & $29.47 \pm 4.89$  & $0.59 \pm 0.19$ \\
                \midrule
                \multirow{2}{*}{\textbf{\shortstack[l]{Oval\\Face}}}     & Biased   & $60.32 \pm 1.20$  & $20.03 \pm 5.55$      & $28.42 \pm 4.38$       & $29.51 \pm 9.60$  & $20.49 \pm 8.19$  & $0.77 \pm 0.23$ \\
                                                                         & Unbiased & $56.51 \pm 3.97$  & $48.84 \pm 2.20$      & $6.58 \pm 2.37$        & $11.24 \pm 7.27$  & $9.39 \pm 6.29$   & $0.66 \pm 0.40$ \\
                \midrule
                \multirow{2}{*}{\textbf{\shortstack[l]{Pale\\Skin}}}     & Biased   & $80.49 \pm 2.03$  & $55.91 \pm 5.39$      & $17.83 \pm 2.69$       & $15.34 \pm 3.84$  & $8.65 \pm 1.92$   & $1.15 \pm 0.17$ \\
                                                                         & Unbiased & $86.96 \pm 0.83$  & $83.37 \pm 1.61$      & $2.99 \pm 0.84$       & $5.47 \pm 2.73$  & $4.94 \pm 2.14$  & $0.81 \pm 0.19$ \\
                \midrule
                \multirow{2}{*}{\textbf{\shortstack[l]{Pointy\\Nose}}}   & Biased   & $63.96 \pm 0.98$  & $31.87 \pm 4.24$      & $22.12 \pm 2.69$       & $27.17 \pm 3.82$  & $22.79 \pm 3.84$  & $1.12 \pm 0.41$ \\
                                                                         & Unbiased & $62.27 \pm 3.09$  & $48.13 \pm 6.07$      & $12.69 \pm 3.29$       & $22.32 \pm 8.46$  & $24.15 \pm 7.24$  & $0.50 \pm 0.19$ \\
                \midrule
                \multirow{2}{*}{\textbf{\shortstack[l]{Straight\\Hair}}} & Biased   & $70.04 \pm 1.46$  & $52.58 \pm 5.27$      & $14.75 \pm 4.82$      & $7.18 \pm 6.37$   & $6.54 \pm 3.02$   & $0.88 \pm 0.31$ \\
                                                                         & Unbiased & $66.92 \pm 1.84$  & $57.43 \pm 3.76$      & $8.73 \pm 2.96$       & $12.05 \pm 6.28$  & $14.77 \pm 6.31$  & $0.65 \pm 0.39$ \\
                \midrule
                \multirow{2}{*}{\textbf{\shortstack[l]{Wearing\\Lipstick}}} & Biased   & $77.79 \pm 0.64$  & $57.05 \pm 1.79$      & $18.73 \pm 0.87$      & $33.87 \pm 2.74$   & $37.20 \pm 1.78$   & $1.01 \pm 0.21$ \\
                                                                            & Unbiased & $73.59 \pm 2.74$  & $57.83 \pm 7.36$      & $14.82 \pm 4.04$      & $24.95 \pm 11.55$  & $28.47 \pm 9.16$   & $0.83 \pm 0.19$ \\
                \midrule
                \multirow{2}{*}{\textbf{Young}}                             & Biased   & $77.21 \pm 0.58$  & $55.16 \pm 3.58$      & $14.09 \pm 1.91$      & $12.59 \pm 2.83$   & $18.29 \pm 2.74$   & $0.95 \pm 0.29$ \\
                                                                            & Unbiased & $71.20 \pm 1.75$  & $62.67 \pm 5.69$      & $7.42 \pm 4.12$       & $14.42 \pm 9.57$   & $13.04 \pm 9.02$   & $0.89 \pm 0.27$ \\
			    \bottomrule
			\end{tabular}
            \end{sc}
		\end{footnotesize}
	\end{center}
        \vspace{-0.275in}
\end{table*}

\begin{figure*}[tb!]
        \centering
        \subfigure[Non-blonde Female]{\includegraphics[width=0.225\textwidth,height=!]{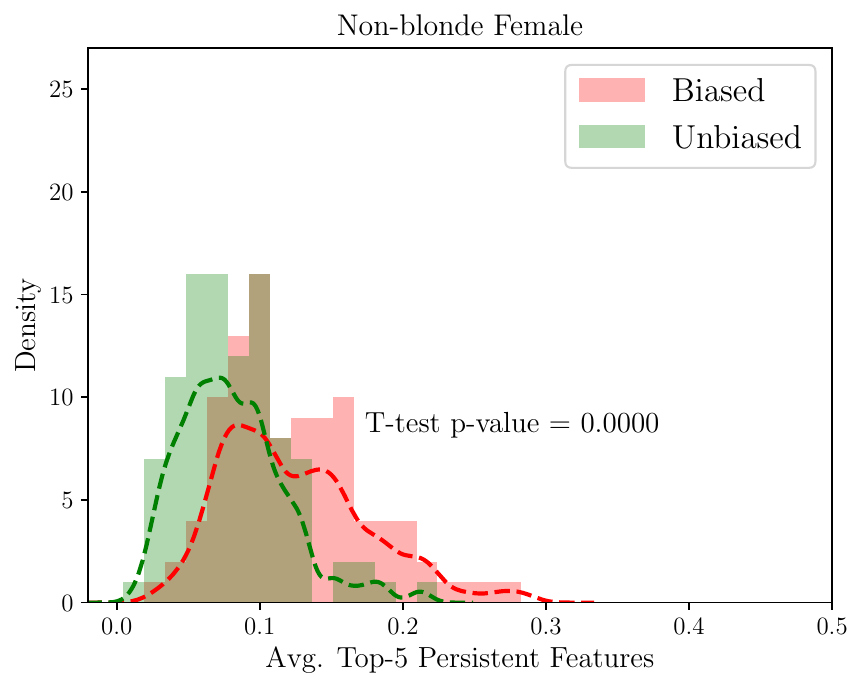}}\hspace{1em}
        \subfigure[Blonde Female]{\includegraphics[width=0.225\textwidth,height=!]{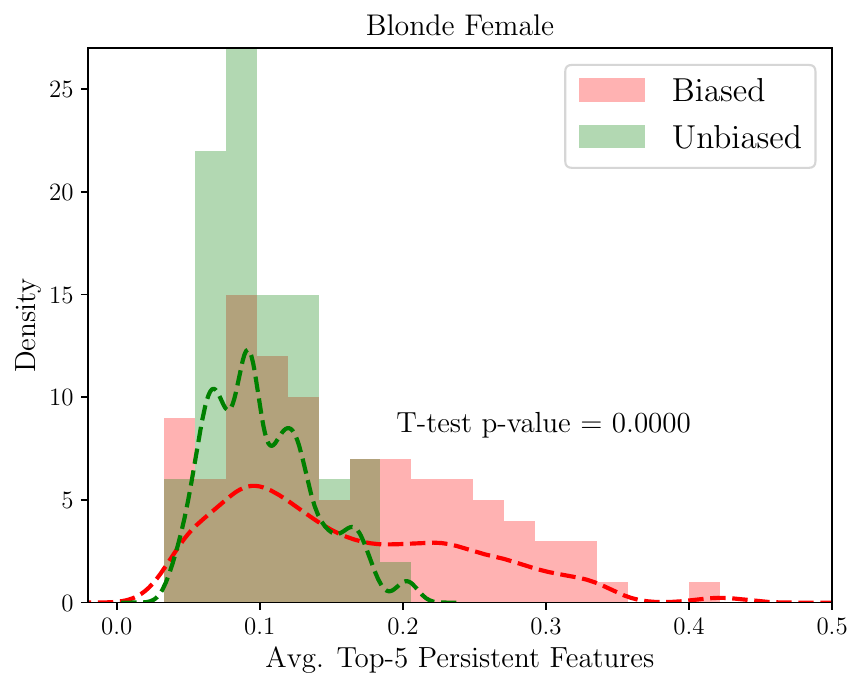}}\hspace{1em}
        \subfigure[Non-blond Male]{\includegraphics[width=0.225\textwidth,height=!]{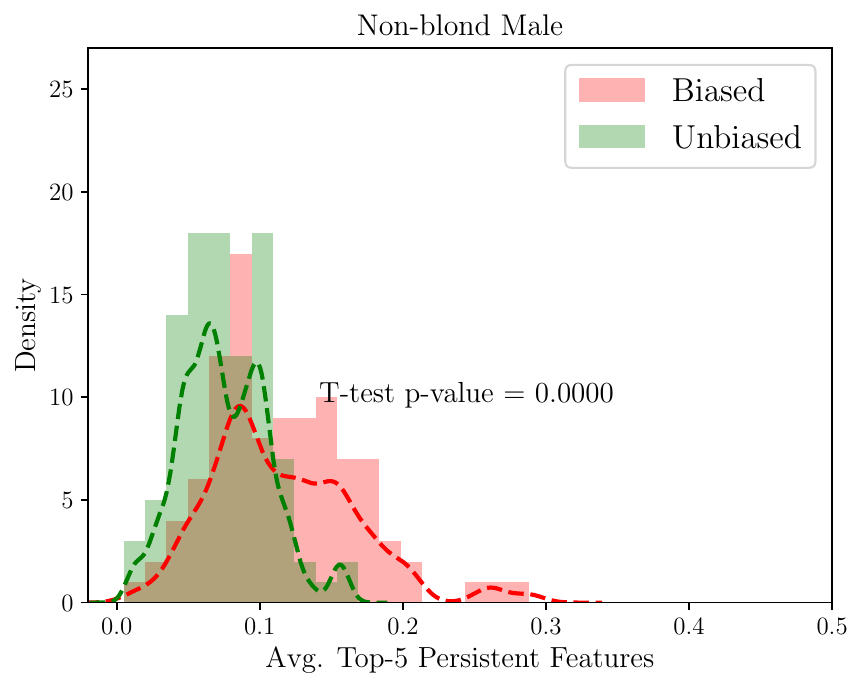}}\hspace{1em}
        \subfigure[Blond Male]{\includegraphics[width=0.225\textwidth,height=!]{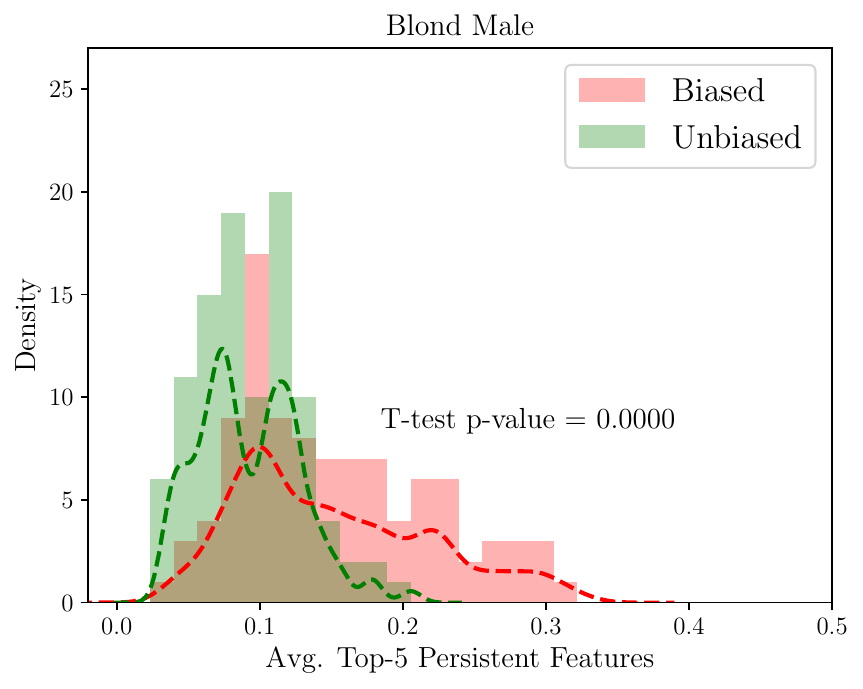}}
     \caption{\label{fig:ttest_fairness} The distribution of average top-5 persistence of 1D homology groups for ResNet-18 models trained with biased and unbiased objectives. Each sub-figure represent a different protected group. The p-value of the T-test has also been shown in each figure.}
     \vspace{0.1in}
\end{figure*}

\paragraph{Settings:}
To conduct our empirical study, we use the state-of-the-art approach of~\citet{seo2022unsupervised} to learn unbiased ResNet-18 models over the CelebA dataset~\citep{liu2015faceattributes}.
Using \textit{gender} as our sensitive attribute, we train DNN models to predict various target attributes such as blonde hair, pale skin, etc.
Naturally, these attributes have a skewed distribution depending on the subject's gender.
For instance, the CelebA dataset contains a higher number of blonde females in comparison to males, making it a great dataset to study fairness in DNNs.
For each target attribute, we train 10 regular (biased) and fair (unbiased) models.
The only exception is the ``Blonde Hair'', for which we train 100 models.

\paragraph{Evaluation:}
To quantitatively evaluate bias in each case, we use common performance metrics such as unbiased accuracy, worst group accuracy, standard deviation of unbiased accuracy over groups, Equalized Odds~(EO), and Average Odds~(AO)~\citep{hardt2023fairness}.
Moreover, we use the framework outlined in~\Cref{sec:framework} to calculate the topological features of each model.
In particular, we compute the 1D persistence diagram of the model and obtain the mean of the top-5 persistent features.

\begin{figure*}[tb!]
        \centering
        \subfigure[\label{fig:corr_fairness_group}]{\includegraphics[width=0.25\textwidth,height=!]{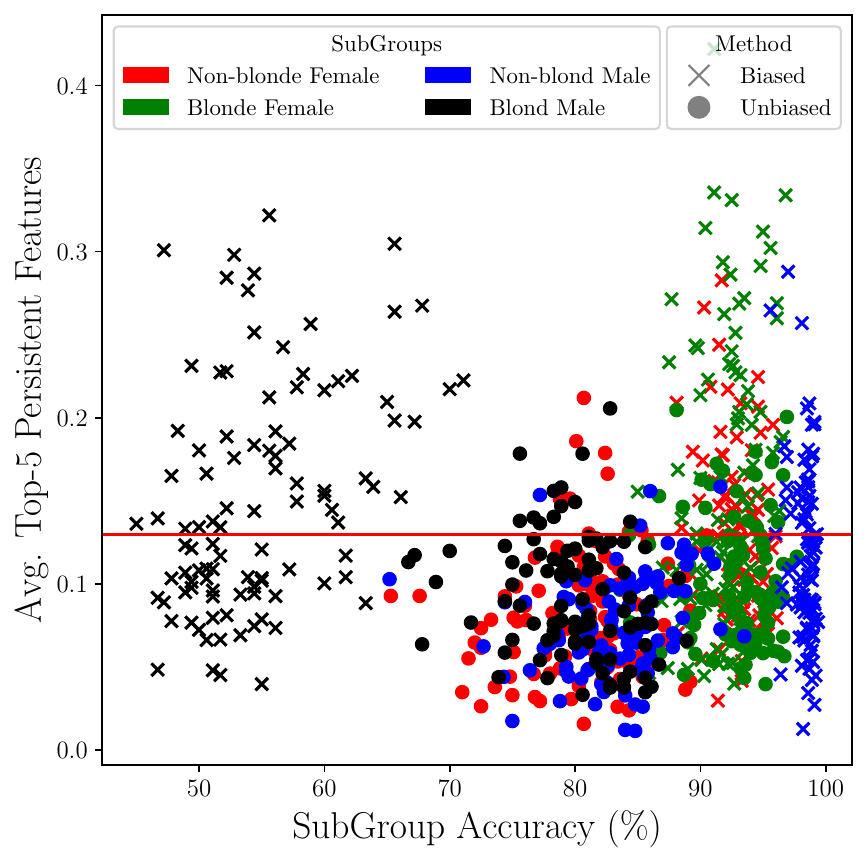}}\hspace{5em}
        \subfigure[\label{fig:corr_fairness_worse_group}]{\includegraphics[width=0.25\textwidth,height=!]{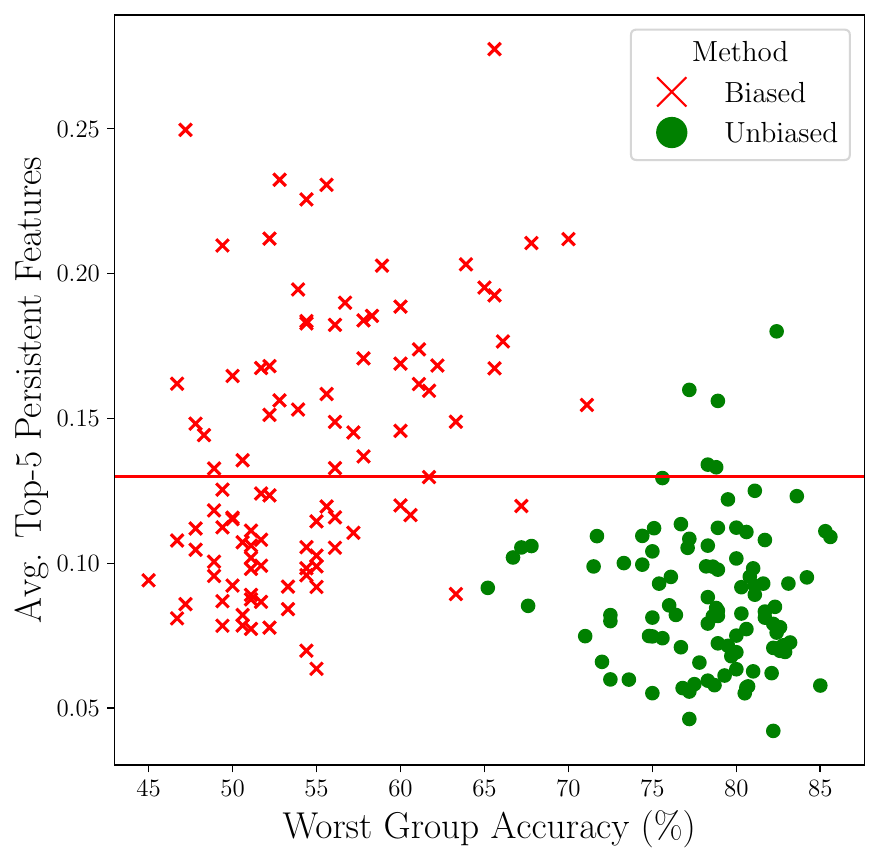}}
     \caption{\label{fig:corr_fairness} The scatter-plot of average top-5 persistence of 1D homology groups against (a) sub-group accuracy and (b) worst-group accuracy. The red line indicates a preset threshold that can separate the biased from unbiased models.}
     \vspace{0.1in}
\end{figure*}

\paragraph{Findings:}
We outline our key findings as below:
\vspace{-0.75em}
\begin{itemize}\setlength\itemsep{0.25em}
    \item As shown in~\Cref{fig:ttest_fairness}, \textbf{biased models exhibit a higher 1D persistence compared to unbiased models}.
    This has been likely caused by the discrepancies in the shortcut trajectories that biased models create within the DNN.
    Again, this difference is statistically significant to distinguish unbiased from biased models.
    \item We can see from~\Cref{fig:corr_fairness} that \textbf{models with higher top-5 persistence are more likely to be biased}.
    This is also evident from the worst group accuracy of these models which are significantly lower than unbiased ones.
    \item In~\Cref{tab:fairness}, we provide a quantitative comparison between biased and unbiased models for 10 target attributes from the CelebA dataset.
    As seen, all \textbf{unbiased models have a lower top-5 persistence compared to their biased counterparts}.
    Interestingly, a lower top-5 persistence signals a lower standard deviation for unbiased accuracy across sensitive groups.
    Therefore, \textbf{there exists a topological signature within DNN models that indicates whether they are biased or unbiased}.
\end{itemize}

\section{Conclusion and Future Directions}
\label{sec:future}
We demonstrated that the trajectories passed by inputs through neural networks contain invaluable insights about shortcuts within the model.
To this end, we used two novel cases of shortcut learning in DNNs: unlearnable examples and fairness.
We saw that persistent homology can play an important role in revealing these failure cases within a neural network.
Interestingly, we showed that these differences are statistically significant.

While we used these two specific case studies, the application of persistent homology to such failure cases is not limited to these cases only.
Previously, PH-based solutions have been independently discovered for other failures of neural networks related to shortcut learning, such as backdoor attacks~\citep{zheng2021topobackdoor}, adversarial examples~\citep{gebhart2017adversary,goibert2022topology}, and out-of-distribution detection~\citep{lacombe2021topouncertainty}.
In this paper, we aimed to argue that all such solutions can be combined to mitigate shortcut learning at a broader level.
This is due to the fact that the cause of most issues with DNNs lie within shortcut learning, which could be traced by exploring the computational graphs trajectories.
Below, we highlight interesting and challenging future directions:
\vspace{-0.5em}
\begin{itemize}\setlength\itemsep{0.25em}
    \item \textbf{Using Persistent Homology to Create Powerful Unlearnable Datasets:} As we see in~\Cref{fig:wasserstein}, the most resilient unlearnable examples are those that deviate the most from the usual trajectories of clean models. Proposing a novel unlearnable example generation objective that reflects such a phenomenon directly could lead to better unlearnable examples.
    \item \textbf{Proposing a New Fairness Measure based on Persistent Homology:} As we saw in~\Cref{tab:fairness}, PH-based measures such as the top-5 persistent features have a tendency towards revealing bias in DNN models. However, these raw measures are not calibrated: even though their relative magnitude could be an indication of bias, we cannot directly compare different models based on this measure. Introducing a new calibrated measure of fairness using persistent homology could be helpful in a better, more universal quantification of fairness in DNN models.
    \item \textbf{Incorporating Topological Measures in Decision-Making:} In a broader sense, quantification of uncertainty using topological measures could become the cornerstone of decision-making using DNNs. This is because apparently normal models should guide the inputs in certain trajectories within the network, and as such, proposing a new measure that covers this could help us detect issues within a neural network in a straightforward manner. Such solutions have also the potential of explainability, as they could reveal the paths involved with a given input during inference.
    \item \textbf{Introducing a Universal Regularizer for DNN Training:} Once we can quantify a universal measure for quantification of uncertainty, we could potentially use these measures to enforce certain behavior within the neural network computational graph. Unfortunately, existing tools for integrating topological measures during neural network training lag behind their counterparts in terms of their computational speed~\citep{hofer2020tdd}, rendering it challenging to use for large-scale models.
    \item \textbf{Testing Solutions under Different Shortcut Learning Scenarios:} Lastly, we encourage researchers in this area to test their solutions not only for a specific type of shortcuts, but for a broad range of them. The creation of a standard benchmark along this line could help researchers to think about the broader issue of shortcut learning rather than focusing on a specific use case.
\end{itemize}


\begin{ack}
Sarah Erfani is in part supported by Australian Research Council~(ARC) Discovery Early Career Researcher Award~(DECRA) DE220100680. 
Moreover, this research was partially supported by the ARC Centre of Excellence for Automated Decision-Making and Society~(CE200100005), and funded partially by the Australian Government through the Australian Research Council.
\end{ack}


\bibliography{sn-bibliography}


\newpage
\appendix
\onecolumn

\section{Extended Discussion on Experimental Results}
In this section, we present an extended version of our experimental results on unlearnable examples in \Cref{sec:cases:unlearnable}.

In \Cref{fig:ttest}, we presented the distribution of average persistence of 1D persistence diagrams~(PDs) for models trained over three unlearnable datasets.
As seen, in all unlearnable models, the population of the topological measure is statistically different from that of clean models.
This signals that inputs in unlearnable models traverse a different trajectory compared to clean models.
To demonstrate this point, we show the top-3 persistent cycles (the ones with the highest life-time) for each unlearnable model along with the clean model in~\Cref{fig:trajectories}:
The trajectories traversed by inputs in the clean model are quite different from those of unlearnable models.
Interestingly, as depicted in~\Cref{fig:features}, these trajectories indicate how the features from input progresses through the model.
In the clean model, we can clearly spot the features of the input image in the intermediate layers of the network, while this is not the case for all the unlearnable models.
This point is also clearly visible from the top-3 trajectories shown in~\Cref{fig:trajectories}.

In \Cref{sec:cases:unlearnable}, we also investigated how topological data analysis can reveal the power of an unlearnable dataset.
In particular, we showed that as the Wasserstein distance between the persistence diagram of the poisoned model and that of the clean model increases, the linear probe accuracy (\textsc{LP Acc}) of that model decreases.
We report similar observations for DenseNet-121~\citep{huang2017densenet} as another DNN architecture.
We use exactly the same settings as the ResNet-18 for this architecture, and train 70 independent models for each unlearnable dataset plus the clean one.
Then, we use the fine-tuning approach of~\citet{segura2023what} to obtain a linear probe accuracy.
\Cref{fig:wasserstein} and \Cref{tab:unlearnable_all} show our results.
As seen, we can make the same conclusions about the Wasserstein distance and its relationship with the power of unlearnable examples for the DenseNet-121 models.

\section{Extra Case Study: Backdoor Attacks}
Backdoor attacks~\citep{gu2017badnets} are a family of data poisoning attacks where the adversary implants a backdoor in the model through poisoning the training data.
This is usually done via attaching certain triggers to some of the training data.
The adversary's goal is to ensure that the poisoned model behaves normally under benign inputs while outputting its intended outcome whenever the trigger is activated.
There have been numerous backdoor attacks such as BadNets~\citep{gu2017badnets}, Blended~\citep{chen2017blended}, PhysicalBA~\citep{li2021physicalba}, Label-Consistence~(LC)~\citep{turner2019label-consistent}, Reflection Backdoor~(Refool)~\citep{liu2020refool}, ISSBA~\citep{li2021issba}, Input-aware Dynamic Backdoors~(IAD)~\citep{nguyen2020IAD}, and WaNet~\citep{nguyen2021wanet} that were proposed to create powerful and stealthy backdoors against neural networks.

It has been commonly believed that backdoor attacks also create a shortcut within the neural network, and that this shortcut is activated whenever the input contains the trigger.
\citet{zheng2021topobackdoor} used this informal belief and devised a topological feature extractor to detect poisoned from benign models.
In this section, we extend our statistical analysis for unlearnable datasets to backdoor attacks.
To this end, we use the~\texttt{BackdoorBox}~\citep{li2023backdoorbox} library\footnote{\url{https://github.com/THUYimingLi/BackdoorBox}} and train 70 ResNet-18 models on the CIFAR-10 dataset for each of the previously mentioned backdoor attacks as well the clean data.
Then, we use the same topological framework of~\citet{zheng2021topobackdoor} to extract the topological features of the trained models.

\begin{figure*}[tb!]
        \centering
        \subfigure[Blended]{\includegraphics[width=0.225\textwidth,height=!]{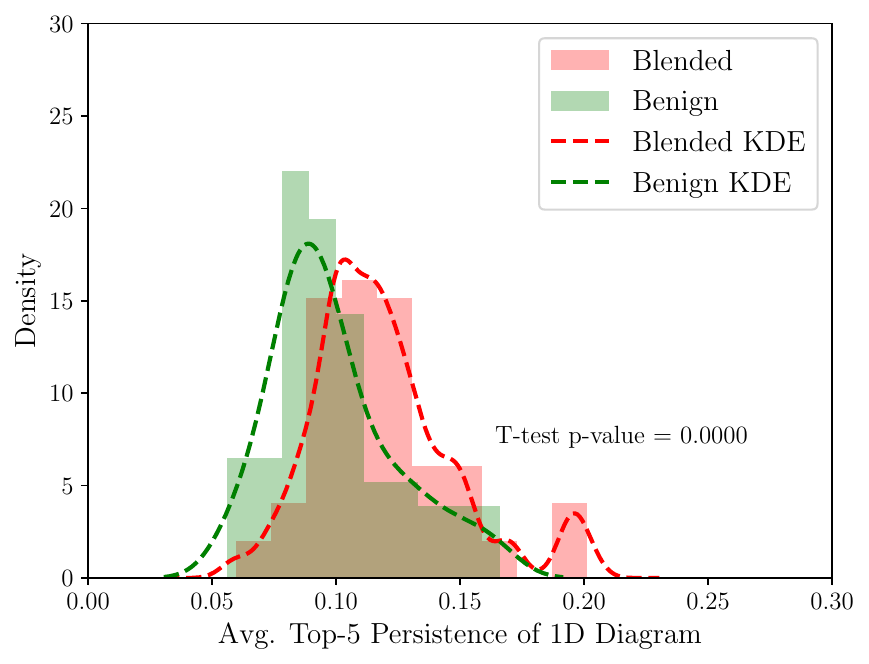}}\hspace{1em}
        \subfigure[IAD]{\includegraphics[width=0.225\textwidth,height=!]{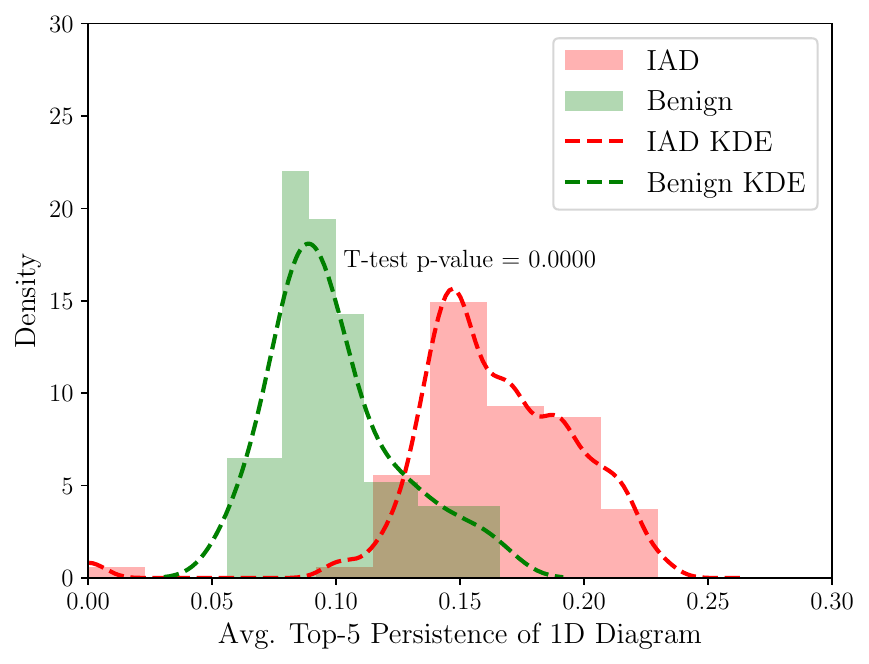}}\hspace{1em}
        \subfigure[PhysicalBA]{\includegraphics[width=0.225\textwidth,height=!]{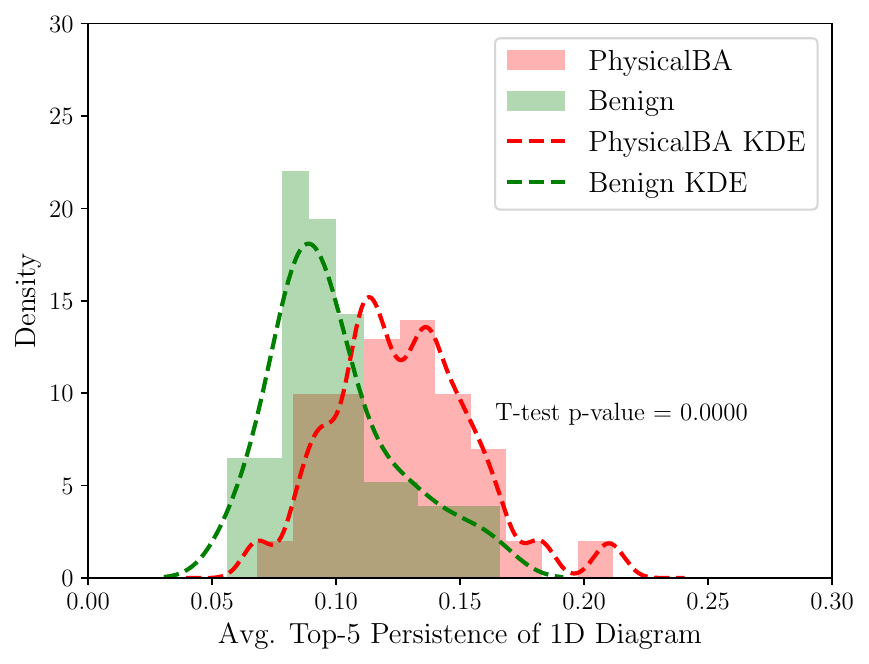}}\hspace{1em}
        \subfigure[Refool]{\includegraphics[width=0.225\textwidth,height=!]{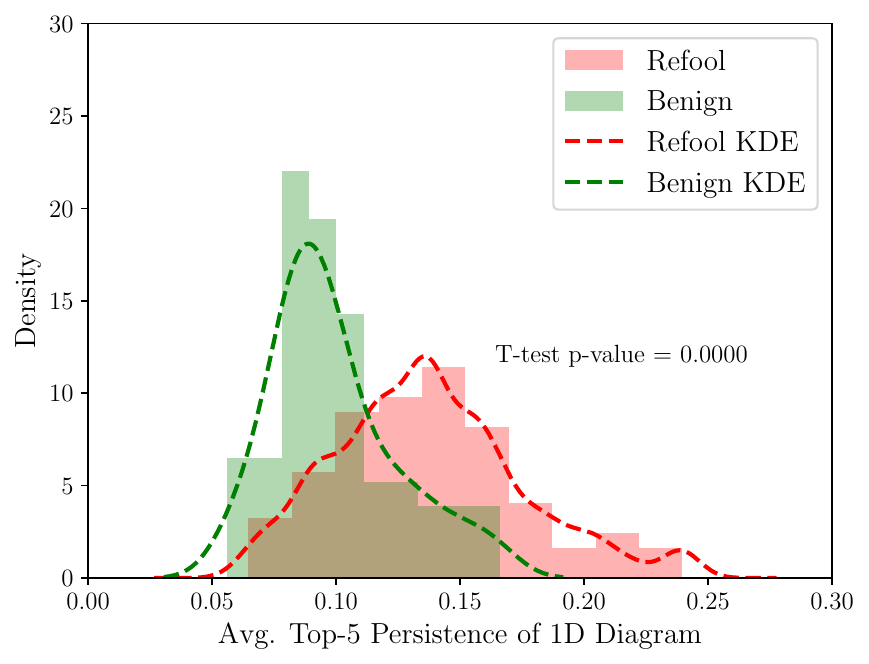}}\\
        \subfigure[BadNets]{\includegraphics[width=0.225\textwidth,height=!]{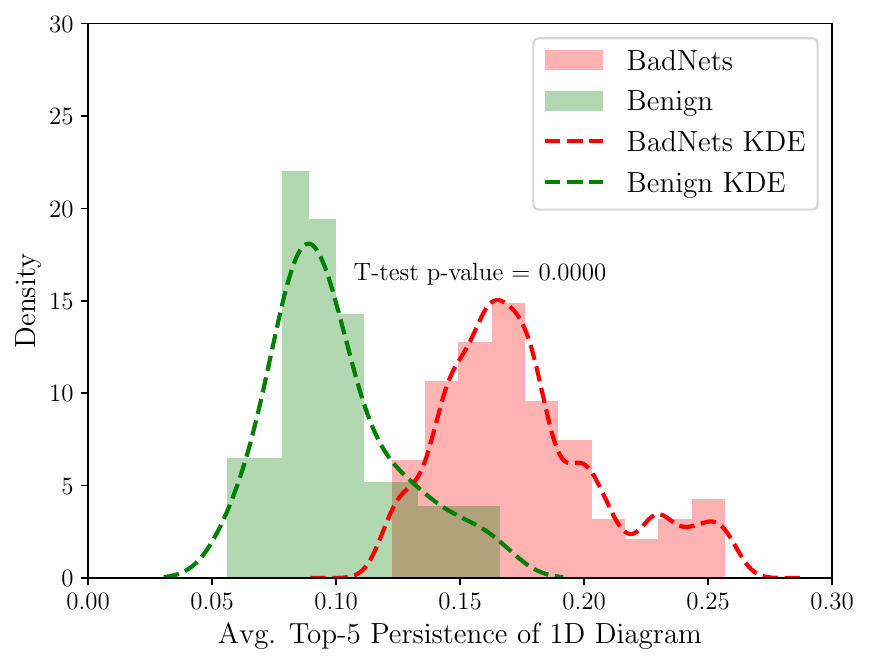}}\hspace{1em}
        \subfigure[WaNet]{\includegraphics[width=0.225\textwidth,height=!]{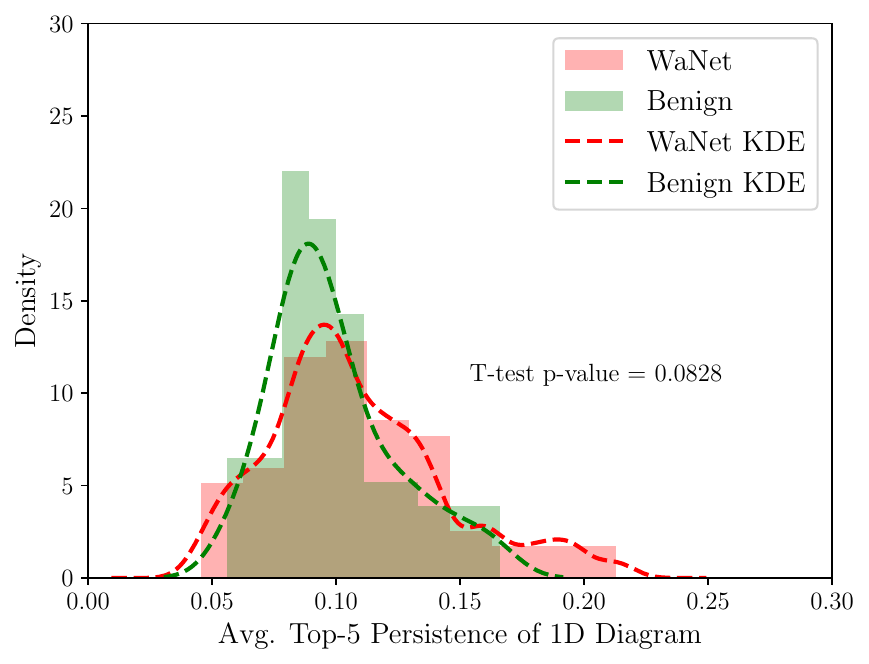}}\hspace{1em}
        \subfigure[LC]{\includegraphics[width=0.225\textwidth,height=!]{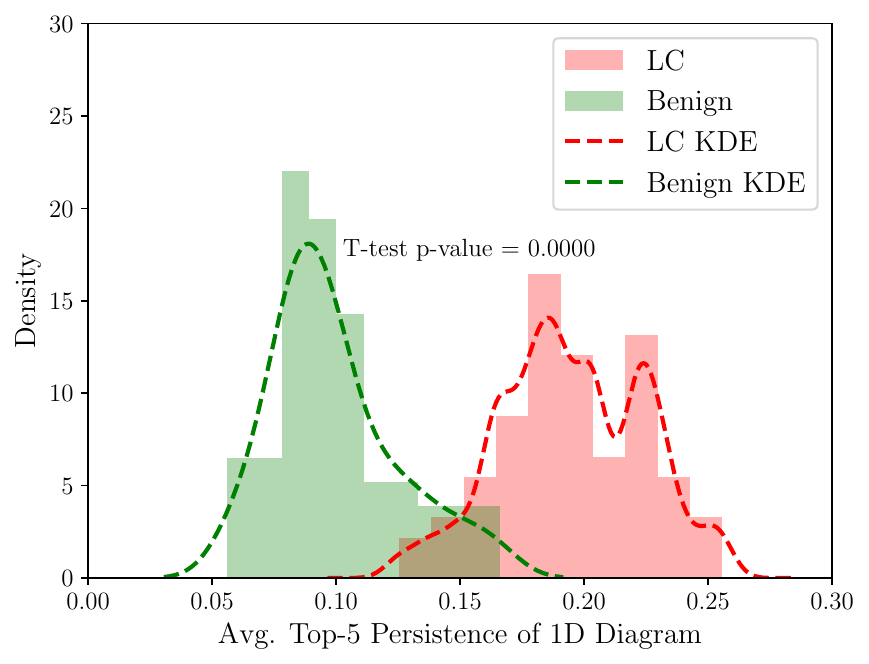}}\hspace{1em}
        \subfigure[ISSBA]{\includegraphics[width=0.225\textwidth,height=!]{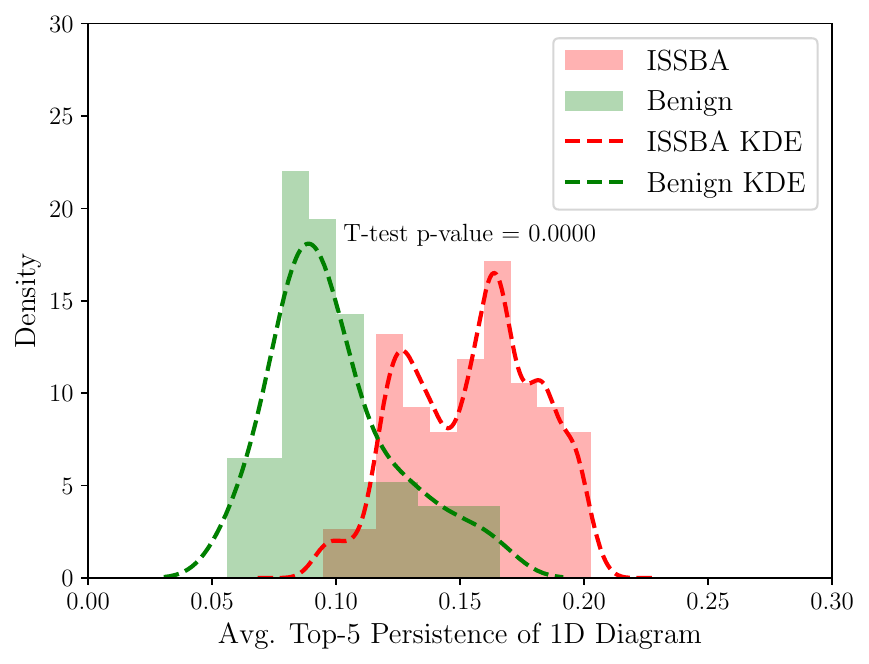}}
     \caption{\label{fig:ttest_backdoor_appendix} The distribution of average top-5 persistence of 1D homology groups for ResNet-18 models trained over various backdoored CIFAR-10 datasets. Each histogram summarizes the result of 70 independent training runs for each dataset. The p-value of the T-test has also been shown in each figure.}
\end{figure*}

\Cref{tab:backdoor_data} and \Cref{fig:ttest_backdoor_appendix} summarize our experimental results on backdoor attacks.
Similar to the previous two case studies, we see that almost all backdoor attacks leave topological signatures within the neural network trajectories and persistent homology can reveal them.
Similar to the other two cases, if we can come up with a differentiable topological regularizer for neural networks, we might be able to mitigate backdoor attacks as well.
As we can see, for all the three case studies, the same topological tool can be applied.
This is because the root cause of all these issues traces back to shortcut learning, which is a malcious behavior in neural networks.

\begin{table*}[p!]
	\caption{Performance of ResNet-18 and DenseNet-121 models trained over data availability attacks on CIFAR-10 dataset. The performance over the clean dataset is shown for reference. The results are averaged over 70 models. \textsc{LP} and \textsc{Avg PD\textsubscript{1}} stand for Linear Probe and Average Persistence of 1D diagram, respectively.}
    \label{tab:unlearnable_all}
	\begin{center}
		\begin{footnotesize}
            \begin{sc}
		    \setlength\tabcolsep{0.5em}
			\def\arraystretch{1.5}
			\begin{tabular}{lcccccc}
				\toprule
                \multirow{2}{*}{\rotatebox{90}{\textbf{Arch}}}
                &\multirow{2}{*}{\textbf{Dataset}}
				&\multicolumn{5}{c}{\textbf{Performance Metrics}}\\
				\cmidrule(lr){3-7}
				&                                  & Train Acc~(\%)   & Test Acc~(\%)    & LP Acc~(\%)      & Avg PD\textsubscript{1}~($\times 10^{-2}$) & WSD\\
				\midrule
                \multirow{7}{*}{\rotatebox{90}{\textbf{ResNet-18}}}
				&\textbf{Clean}                    & $97.44 \pm 0.00$ & $86.78 \pm 0.25$ & $86.51 \pm 0.28$ & $2.19 \pm 0.50$ & $0.44 \pm 0.10$\\
                &\textbf{TAP}                      & $100.0 \pm 0.00$ & $17.71 \pm 1.81$ & $57.96 \pm 0.90$ & $2.84 \pm 0.60$ & $0.61 \pm 0.08$\\
                &\textbf{SHR}                      & $100.0 \pm 0.00$ & $12.59 \pm 1.55$ & $42.77 \pm 0.91$ & $3.10 \pm 0.69$ & $0.92 \pm 0.21$\\
                &\textbf{REM}                      & $100.0 \pm 0.00$ & $21.95 \pm 1.52$ & $35.77 \pm 0.84$ & $3.80 \pm 0.77$ & $1.10 \pm 0.16$\\
                &\textbf{EMN}                      & $100.0 \pm 0.00$ & $20.83 \pm 0.40$ & $33.05 \pm 0.65$ & $4.39 \pm 0.90$ & $1.02 \pm 0.20$\\
                &\textbf{NTGA}                     & $100.0 \pm 0.00$ & $12.69 \pm 2.42$ & $31.00 \pm 2.51$ & $3.32 \pm 0.62$ & $0.97 \pm 0.25$\\
                &\textbf{AR}                       & $100.0 \pm 0.00$ & $13.21 \pm 1.37$ & $27.58 \pm 1.50$ & $3.86 \pm 0.67$ & $1.31 \pm 0.17$\\
                \midrule
                \multirow{7}{*}{\rotatebox{90}{\textbf{DenseNet-121}}}
				&\textbf{Clean}                    & $97.44 \pm 0.00$ & $95.05 \pm 0.15$ & $94.93 \pm 0.15$ & $2.70 \pm 0.30$ & $1.98 \pm 0.30$\\
                &\textbf{TAP}                      & $100.0 \pm 0.00$ & $13.13 \pm 1.41$ & $72.10 \pm 0.49$ & $2.38 \pm 0.26$ & $2.16 \pm 0.28$\\
                &\textbf{EMN}                      & $100.0 \pm 0.00$ & $24.15 \pm 0.92$ & $54.28 \pm 1.55$ & $2.41 \pm 0.24$ & $2.61 \pm 0.33$\\
                &\textbf{NTGA}                     & $100.0 \pm 0.00$ & $13.74 \pm 4.19$ & $51.88 \pm 3.37$ & $2.51 \pm 0.29$ & $3.11 \pm 0.48$\\
                &\textbf{SHR}                      & $100.0 \pm 0.00$ & $15.33 \pm 1.33$ & $50.29 \pm 1.07$ & $2.51 \pm 0.29$ & $3.11 \pm 0.39$\\
                &\textbf{REM}                      & $100.0 \pm 0.00$ & $20.34 \pm 1.57$ & $43.29 \pm 0.78$ & $2.89 \pm 0.18$ & $4.69 \pm 0.46$\\
                &\textbf{AR}                       & $100.0 \pm 0.00$ & $13.23 \pm 1.49$ & $32.55 \pm 1.98$ & $2.96 \pm 0.30$ & $4.75 \pm 0.58$\\
			    \bottomrule
			\end{tabular}
            \end{sc}
		\end{footnotesize}
	\end{center}
\end{table*}

\begin{table*}[p!]
	\caption{Performance of ResNet-18 models trained over various backdoor attacks on CIFAR-10 dataset. The performance over the benign dataset is shown for reference. The results are averaged over 70 models.}
    \label{tab:backdoor_data}
	\begin{center}
		\begin{footnotesize}
            \begin{sc}
		    \setlength\tabcolsep{0.5em}
			\def\arraystretch{1.5}
			\begin{tabular}{lcccc}
				\toprule
                \multirow{3}{*}{\rotatebox{90}{\textbf{Arch}}}
                &\multirow{2}{*}{\textbf{Attack}}
				&\multicolumn{3}{c}{\textbf{Performance Metrics}}\\
				\cmidrule(lr){3-5}
				&                                  & \multirow{1}{*}{Test Acc~(\%)}    & \multirow{1}{*}{Backdoor Acc~(\%)} & Avg Top-5 PD\textsubscript{1}~($\times 10^{-1}$) \\
				\midrule
                \multirow{9}{*}{\rotatebox{90}{\textbf{ResNet-18}}}
				&\textbf{Benign}                   & $91.49 \pm 0.29$ & $-$          & $0.99 \pm 0.25$\\
                &\textbf{WaNet}                    & $90.42 \pm 0.25$ & $98.62 \pm 0.73$ & $1.08 \pm 0.36$\\
                &\textbf{Blended}                  & $90.55 \pm 0.32$ & $83.84 \pm 0.68$ & $1.19 \pm 0.29$\\
                &\textbf{PhysicalBA}               & $93.54 \pm 0.22$ & $96.56 \pm 0.32$ & $1.27 \pm 0.29$\\
                &\textbf{Refool}                   & $90.98 \pm 0.29$ & $96.68 \pm 0.44$ & $1.38 \pm 0.38$\\
                &\textbf{ISSBA}                    & $94.36 \pm 0.17$ & $99.85 \pm 0.35$ & $1.55 \pm 0.26$\\
                &\textbf{IAD}                      & $93.88 \pm 0.16$ & $96.63 \pm 1.37$ & $1.64 \pm 0.34$\\
                &\textbf{BadNets}                  & $91.17 \pm 0.26$ & $97.29 \pm 0.11$ & $1.76 \pm 0.32$\\
                &\textbf{LC}                       & $91.53 \pm 0.30$ & $99.80 \pm 0.38$ & $1.95 \pm 0.29$\\
			    \bottomrule
			\end{tabular}
            \end{sc}
		\end{footnotesize}
	\end{center}
\end{table*}

\end{document}